\documentclass[11pt]{article}

\usepackage[preprint]{acl}

\usepackage{times}
\usepackage{latexsym}

\usepackage[T1]{fontenc}

\usepackage[utf8]{inputenc}

\usepackage{microtype}

\usepackage{inconsolata}

\usepackage{graphicx}
\usepackage[english]{babel}
\usepackage[english=british]{csquotes}
\usepackage{tcolorbox}
\usepackage{enumerate}
\usepackage{booktabs}
\usepackage{multirow}
\usepackage{listings}

\title{Beyond the pale: Assessing prevalence and contents\\ of extremist speech in LLM training data}

\author{Dmitry Nikolaev \quad Ashley A.\ Mattheis \\
        University of Manchester \\ \texttt{\{dmitry.nikolaev,ashley.mattheis\}@manchester.ac.uk}}

\begin{document}
\maketitle
\begin{abstract}

Despite a strong interest on the part of the research community in the topic of trustworthy and safe AI, the composition of the text corpora that large language models (LLMs) encounter in pre- and post-training has not yet drawn much attention. In this work, we address the question of whether LLMs are exposed to unfiltered, uncontextualised extremist speech. Using several definitions of extremist speech, stemming from official documents and research literature, and an extraction pipeline combining automated text processing with expert verification, we provide a lower bound on the prevalence of extremist documents in Dolma, an open training corpus underpinning the OLMo series of models. We show that Dolma is likely to include hundreds of thousands of documents containing extremist content and hate speech of several types, including direct calls for violence, and discuss the implications of this for data curation and model pre-training.

\end{abstract}

\section{Introduction}

The subject of trustworthy and safe AI has become one of the largest foci of research interest and grant support in the area of AI/machine learning \citep{kaur2022trustworthy,li2023trustworthy,ferdaus2026towards}. The focus of this research activity seems to be largely on the architectural developments and post-training regimes. In contrast, the problem of data curation and filtering, while being acknowledged \citep[e.g.,][]{liang2022advances}, does not receive nearly as much attention. One particular manifestation of this imbalance is that while there is a healthy body of work analysing political biases and leanings of LLMs \citep[cf.\,][among many others]{motoki2024more,ceron2024beyond,aksoy2026evaluating,nikolaev2026llmsreadyhardchoices}, only a handful of studies have tried to connect them with the composition of the data used at the pre- and post-training stages \citep{stammbach-etal-2024-aligning,xu-etal-2025-better,ceron2026politicalcontentllmspre}.

One area where the topic of trustworthy and safe AI strongly overlaps with the purview of computational social science is identification and analysis of extremist speech. It has been repeatedly observed that due to data scarcity automated identification of fringe political positions is a hard task for LLMs \citep{nikolaev-etal-2023-multilingual,bol2025can}. At the same time, even a small amount of extremist content in the training data, due to LLMs' strong capacity for memorising texts verbatim, can lead to undesired exposure of users, including underage ones, to dangerous viewpoints. Worse still this can lead to gradual normalisation of such viewpoints if their proponents and key texts expounding them are mentioned without proper commentary in models' outputs. Some worrying trends in LLMs' relations to extremist or intolerant ideologies have been reported over the last several years \citep{mcguffie2020radicalizationrisksgpt3advanced,Abid2021,Lakomy2023,baele2024}.

\begin{figure}[t]
    \centering
    \includegraphics[width=1\linewidth]{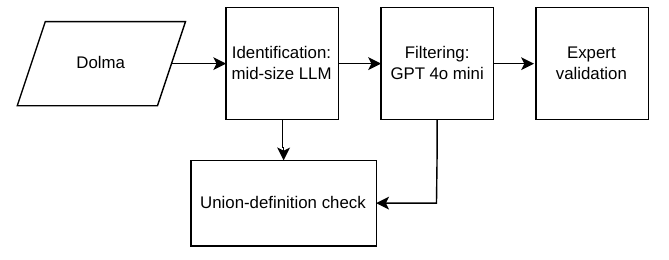}
    \caption{The analytic pipeline implemented in the study.}
    \label{fig:pipeline}
\end{figure}

It is an open question whether extremist speech should be as thoroughly as possible excised from the training data to eliminate the risks outlined above or whether it should be retained with proper contextualisation. As a first step, however, it is necessary to at least estimate its prevalence in the training corpora. This is what we aim to do in this study. By combining large-scale automated classification and filtering with manual expert analysis and targeting Dolma \citep{soldaini-etal-2024-dolma}, one the largest and most authoritative open-source pre-training datasets, we show that extremist content of several types is indeed present in the training data of LLMs.

Our methodological contribution is a conservative pipeline for identifying such texts, which provides a reliable lower bound on their prevalence. We further discuss the dependence of the results on the precise definition of extremism used for model prompting. Each of the three definitions that we tested leads to slightly different outcomes that may be desirable in different practical applications.

The paper is organised as follows: in Section~\ref{sec:definitions}, we present the definitions of extremism that we used for identifying extremist and hateful speech.
In Section~\ref{sec:data-and-methods}, we present our experimental setup. The prevalence estimates and robustness checks are reported in Section~\ref{sec:prevalence}. Section~\ref{sec:analysis} presents a in-depth analysis of a sample of documents extracted using different definitions.
Section~\ref{sec:source_domains} surveys the sources of the manually-verified extremist documents as reflected in source-domain annotations in Dolma,
and Section~\ref{sec:conclusion} concludes.

\section{Extremism definitions}
\label{sec:definitions}

The status of extremism and extremist speech differs across countries. For example, in the US, with its strong First Amendment protections, extremist speech is not proscribed as such and is mostly analysed in its connections with actual or potential terrorist activities \citep{Tsomidis16032022}. At the same time, in the UK, while extremist views and speech are not directly proscribed either, the government takes it upon itself to avoid \enquote{providing a platform, funding or legitimacy to individuals, groups or organisations who attempt to advance extremist ideologies} \citep{uk2024extremism}, which necessitates having official guidance on what such ideologies are. In our analysis, we use the official UK definition of extremism, together with two definitions provided by researchers working in the field of extremism and terrorism studies.

\subsection{The UK definition}

The following definition was minimally adapted from \citet{uk2024extremism} to make it self contained:

\begin{quote}
Extremism is the promotion or advancement of an ideology based on violence, hatred or intolerance, that aims to:

\begin{enumerate}
    \item negate or destroy the fundamental rights and freedoms of others; or
    \item undermine, overturn or replace the system of liberal democracy and democratic rights; or
    \item intentionally create a permissive environment for others to achieve the results in (1) or (2).
\end{enumerate}
\end{quote}
%
This definition is focused on the opposition of extremist actors and speakers to the democratic order and the system of rights and freedoms prevailing in developed societies.

\subsection{Berger's definition}

The following definition was adapted from \citet[44]{berger2018}. This definition is focused on the somewhat abstract opposition of the in-group and the out-group, which places some interpretative burden on the analyst using it:

\begin{quote}
Extremism refers to the belief that the success or survival of the speaker’s or author’s in-group can never be separated from the need for hostile action against an out-group. The hostile action must be part of the in-group’s definition of success. Hostile acts can range from verbal attacks and diminishment to discriminatory behavior, violence, and even genocide.    
\end{quote}

\subsection{Schmid's definition}

This definition has been adapted from \citet[630]{schmid2013}. It is rather detailed and contains a list of concrete traits and ideological positions characteristic of extremist actors:

\begin{quote}
In the context of democratic societies, extremism is a form of political expression (usually on the far left or the far right of the political spectrum) that is not acceptable to the more moderate mainstream of political life. Extremist groups and parties tend to be anti-constitutional, antidemocratic, anti-pluralistic, fanatical, intolerant, non-compromising, single-minded, authoritarian and adhering to an ends-justify-means philosophy, wanting to realise their goals by any means, including the use of political violence against opponents. Extremists on the political left or right and those of a religious-fundamentalist orientation favour violence over persuasion, uniformity over diversity, unity over pluralism and orders over dialogue.
\end{quote}

\section{Data and methods}
\label{sec:data-and-methods}

Following the approach suggested by \citet{ceron2026politicalcontentllmspre}, we target the Dolma corpus \citep{soldaini-etal-2024-dolma}, a fully-open and well-documented pre-training corpus. It has been compiled for training the OLMo series of models \citep{olmo20242olmo2furious} and is not perfectly representative of the training data used in other LLMs, especially those based on proprietary corpora. However, the bulk of its contents comes from commonly-used datasets such as CommonCrawl and C4, and it is highly probable that documents found in Dolma have been ingested by most general-purpose models in use today.

The full corpus consists of approximately 3~trillion tokens of English text, which makes a wholistic analysis of it impossible. Therefore, again following \citet{ceron2026politicalcontentllmspre}, we resort to reservoir sampling and use a random sample of 200k documents extracted over one pass over the whole dataset. We then implement a multi-stage analytical pipeline for processing the sample. The pipeline is summarised in Figure~\ref{fig:pipeline}.

\paragraph{Automated extraction.} The first stage consists of classifying the documents as examples of extremist speech or otherwise using several mid-size LMs, with parameter count and the level of quantisation selected in such a way as to make it possible to process the documents in the sample using a single 80GB A100 GPU or two 40GB A100 GPUs. We contrast two models that are the biggest in this \enquote{weight category} (Llama 3.3 70B and Qwen 2.5 72B, both with 4-bit quantisation) with two smaller but more modern models (Qwen 3.5 27B and Gemma 4 31B, both with 8-bit quantisation). The task specification and the extremism definition are provided in the system prompt, with the target text provided in the user prompt. Models were prompted in the zero-shot fashion.

\paragraph{Union check.} In order to verify that the models have a general understanding of the task, in addition to prompting them with individual definitions, we also use a union prompt where definitions are given together and the model is asked to decide if the input satisfies any of them. We expect a model that is well-suited to the task to 
\begin{enumerate}[i]
    \item return more texts for the union definition than for each of the individual definitions and
    \item to have a reasonable overlap between the sample extracted using the union definition and the samples extracted using individual definitions.
\end{enumerate}
E.g., if a model returns 10 texts using the UK definition and 100 texts using the union definition, we hope to see that the intersection between these two samples has at least 7 texts, otherwise the LLMs' understanding of the prompts would be very questionable.

The single-definition and union prompts can be found in \ref{sec:appendix-prompts}.

\paragraph{Automated filtering.} If a model passes the union check, we may assume that it is following the prompt in a more or less consistent manner. However, consistency by itself does not safeguard against the major types of extraction errors, i.e.\ false positives and false negatives.

Unfortunately, we do not have a convincing way for screening for false negatives. We cannot recheck all discarded documents for practical reasons, and a smaller random sample is unlikely to provide a good estimate because the actual prevalence of extremist documents is hopefully low. Therefore we concede defeat here and treat our estimate as the lower bound on the prevalence of extremist materials in the training data.

As for false positives, we first test for them by rechecking all extracted documents using GPT 4o mini, which has a good track record on open-domain classification tasks \citep{Roumeliotis2024,Chae2025,Vasudevan2026}, accessed through the OpenAI API. We only used the union definition at this stage.

\paragraph{Expert evaluation.} As the final step, we sampled ten texts per definition from those extracted by one of the models (Gemma~4) and validated by GPT 4o mini. These texts were then analysed by the second author, who is a domain expert in extremist speech. The sample is small, but the extracted texts, given in \ref{sec:appendix-sample}, are long, not entirely coherent, and written in a mixture of hateful and coded language, which makes them very hard to work with. The results of this analysis are presented in Section~\ref{sec:analysis}.

\section{Prevalence estimates}
\label{sec:prevalence}

\begin{table*}[htbp]
\centering
\begin{tabular}{lcccc l}
\toprule
 & Berger & Schmid & UK & Union & Model \\
\midrule
Berger & 1356 (286) & 687 (228)  & 525 (220) & 906 (256)  & \multirow{4}{*}{Llama 3} \\
Schmid &            & 1058 (241) & 499 (206) & 799 (231)  & \\
UK     &            &            & 587 (225) & 557 (216)  & \\
Union  &            &            &           & 1262 (280) & \\
\midrule
Berger & 373 (189)  & 316 (179)  & 309 (177) & 359 (188)  & \multirow{4}{*}{Qwen 2.5} \\
Schmid &            & 680 (233)  & 457 (213) & 655 (233)  & \\
UK     &            &            & 519 (221) & 510 (221)  & \\
Union  &            &            &           & 986 (274)  & \\
\midrule
Berger & 367 (208)  & 252 (177)  & 258 (178) & 312 (189)  & \multirow{4}{*}{Qwen 3.5} \\
Schmid &            & 329 (193)  & 273 (180) & 305 (179)  & \\
UK     &            &            & 315 (195) & 293 (180)  & \\
Union  &            &            &           & 526 (241)  & \\
\midrule
Berger & 301 (176)  & 155 (133)  & 153 (128) & 232 (162)  & \multirow{4}{*}{Gemma~4} \\
Schmid &            & 205 (157)  & 154 (135) & 189 (152)  & \\
UK     &            &            & 195 (150) & 189 (149)  & \\
Union  &            &            &           & 390 (217)  & \\
\bottomrule
\end{tabular}
\caption{Overlaps between the text sets extracted from the 200k Dolma sample based on different definitions. Numbers in parentheses are texts validated as extremist by GPT 4o mini.}
\label{tab:model_comparisons}
\end{table*}

The estimates of the prevalence of extremist texts in Dolma are shown in Table~\ref{tab:model_comparisons}. The first obvious result is that Llama~3 70B is inconsistent and cannot be relied upon: it extracted more texts for a single definition (Berger's) than for the union definition, the results of union checks (i.e.\ differences between the values on the main diagonal and the last column) are very bad, and GPT 4o mini disqualified the lion's share of its results.

The Qwen models, despite vast differences in the model size (72B vs.\ 27B parameters) demonstrate similar patterns, with Qwen~3.5 being much more conservative and better aligned with GPT 4o mini estimates. Interestingly, while Qwen~2.5 seems to return considerably more false positives, it has good results on the union checks, indicating a high degree of internal consistency.

Finally, Gemma~4 has the lowest percentage of extracted texts struck down by GPT 4o mini. This, however, comes at a cost of reduced recall (20\% fewer texts returned on average across definitions compared to Qwen~3.5), and the union-check results are worse (the discrepancy between the sizes of single-definition sets and corresponding overlap-with-union sets is only 3\% on average for Qwen~3.5 but 11\% for Gemma).

These results seem to indicate that Qwen 3.5 27B provides the best balance between internal consistency and precision, as long as the texts extracted using it can be further screened by a larger model or a human expert. Its recall is not noticeably lower on average than that of Llama 3 70B, which seems to have an extremely high false-positive rate. At the same time, Gemma~4 coupled with Schmid's and the UK definitions provides most conservative and reliable results out of the box.

Another conclusion that can be made is that Berger's definition, with its focus on the opposition between the \enquote{in-group} and the \enquote{out-group}, predictably leads to high variance in the estimates. This makes it more suitable for exploratory work but probably less suitable for practical screening applications. As for the other two definitions, the one suggested by Schmid returns more texts than the official UK definition; however, the difference becomes much smaller after GPT 4o mini filtering. Moreover the overlap set is very close in its content to both single-definition sets: the average difference between the two Qwen models and Gemma is only 8.4\%. In practical quantitative terms, therefore, despite marked differences in wording and the level of detail, Schmid's and the UK definition are almost equivalent. See more discussion of the differences between texts extracted using different definitions in Section~\ref{sec:analysis}.

\section{In-depth analysis}
\label{sec:analysis}

In this section, we first briefly outline our approach to coding and analysing the sampled texts (\S~\ref{ssec:coding}) and then survey the results by definition (\S~\ref{ssec:modelling-outcomes}).

\subsection{Coding extremist content in the sample}
\label{ssec:coding}

\begin{table}[t]
\centering
\begin{tabular}{@{}ll@{}}
\toprule
Code & Content   Type \\ \midrule
HS & \begin{tabular}[c]{@{}l@{}}Hate speech, no extremist \\ narratives present\end{tabular} \\
LWX & \begin{tabular}[c]{@{}l@{}}Left-wing extremist narratives are\\ present\end{tabular} \\
MAIN & Mainstreaming of extreme beliefs \\
News & News articles \\
OOK & \begin{tabular}[c]{@{}l@{}}Other texts no extremism / \\ okay for training data\end{tabular} \\
OPR & \begin{tabular}[c]{@{}l@{}}Other texts no extremism / \\ problematic for training data\end{tabular} \\
UKNCON & \begin{tabular}[c]{@{}l@{}}Unknown context with respect to \\ extremism\end{tabular} \\
XRW & \begin{tabular}[c]{@{}l@{}}Right-wing extremist narratives \\ are present\end{tabular} \\
XRW-V & \begin{tabular}[c]{@{}l@{}}Violent right-wing extremist\\ narratives are present\end{tabular} \\ \bottomrule
\end{tabular}
\caption{Codes for in-depth sample analysis.}
\label{tab:codes}
\end{table}

To further assess the validity of the classification and filtering processes, 30 sample texts, 10 from each definitional model, were randomly selected for manual, close reading. This in-depth analysis provides insight into how well the models identified extremist texts from the training data as well as an initial look at what range of text types each definitional model is classifying.

Across the sample, text lengths ranged from a single paragraph to approximately four pages each. In most cases they were single author texts; however, there were a few conversational threads with multiple authors with no post delineations or line breaks such that they looked like single texts.

The codes for the sample texts were generated using a grounded interpretive approach and thus derived from the sample texts themselves. Three close readings were performed on different days to derive the codes, assess the content, and review and verify the coding selections. Coding for extremist speech relied on the inclusion of at least one of the following:
\begin{enumerate}
    \item Known extremist narratives;
    \item Promotion of extremist actors or authors;
    \item Repetition of extremist conspiracy theories \citep[2--18]{waltman2017}.
\end{enumerate}
To complete the coding, 10 codes were identified to categorise the sample texts (as shown in Table~\ref{tab:codes}).

Each of the models does identify extremist content that falls in line with its respective definition. However, each model also identified texts that the in-depth analysis determined were not extremist. The types of texts that each model classified and filtered varied based on which definition was used. Each are discussed in more detail below. 

Interestingly, none of the models returned Jihadist extremist narratives, and only one news story about Jihadist terrorists was returned across the full sample. This may be an effect of technology platform compliance with strict regulation using international or corporate consensus for the proscription of terrorist organisations (e.g., designation lists) which largely focus on Jihadism \citep{borelli2023}.
This discrepancy may also simply highlight the prevalence of content from the specific sources where that particular LLM training data were scraped.

\subsection{Modelling outcomes}
\label{ssec:modelling-outcomes}

Of the 30 sample texts we reviewed, the most prevalent type were 10 (33.34\%) texts classified under close reading as extremist content. These included nine texts predominantly made up of right-wing extremist narratives (XRW), of which two were classified as explicitly calling for violence (XRW-V). A~single text was classified as left-wing extremist (XLW).

\begin{figure}
    \centering
    \includegraphics[width=\linewidth]{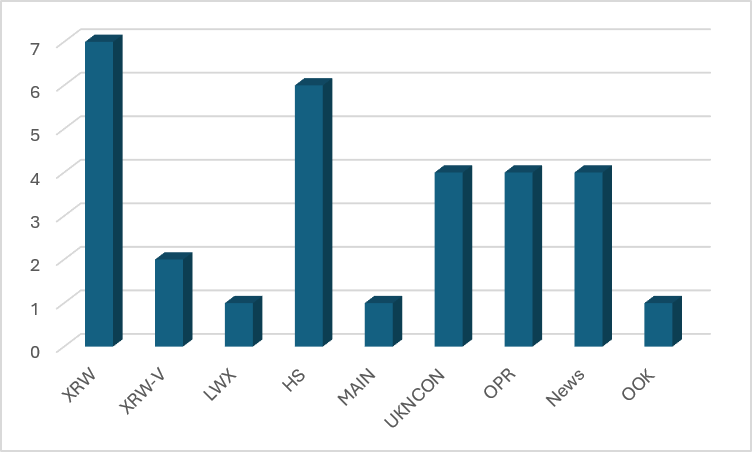}
    \caption{Coding across the total sample.}
    \label{fig:coding}
\end{figure}

The second most common type included 6 (20\%) texts classified as hate speech, meaning that they expressed clear hatred of an out-group (anti-Islam, anti-Semitism, racial hatred, homophobia, etc.) but did not include common extremist narratives or content within the text. The third most prevalent type, made up of 4 (13.34\%) texts each, was split between three non-extremist classifications including:

\begin{enumerate}
    \item News articles (News) that were about extremist or terrorist events, which is likely why they were identified by the model;
    \item Other texts problematic for training data (OPR) which included highly partisan, polarised, or violent content that did not incorporate extremist narratives or content but were extreme, which is likely why they were identified by the model; and
    \item Unknown Context (UNKCON) which included locally specific seemingly political discussions from outside the Anglo-European context and were thus unclear in relation to extremism without further context. Again these were likely returned by the models due to their extreme, partisan, and hateful language.
\end{enumerate}
Additionally, there were two single texts (3.3\%) that were categorised respectively as 1) not extremist and unproblematic in training data (OOK) and 2) mainstreaming of extreme ideas (MAIN). The text classified MAIN is important because its aim is to normalise extremist ideas and such texts are often designed not to be violative, and as such highlights the issue of so-called \enquote{borderline} content \citep{Macdonald2023} and issues around how such narratives that cannot be clearly classified as extremist should be addressed. Figure~\ref{fig:coding} shows the spread of codes for the whole sample.

\begin{figure}[t]
    \centering
    \includegraphics[width=1\linewidth]{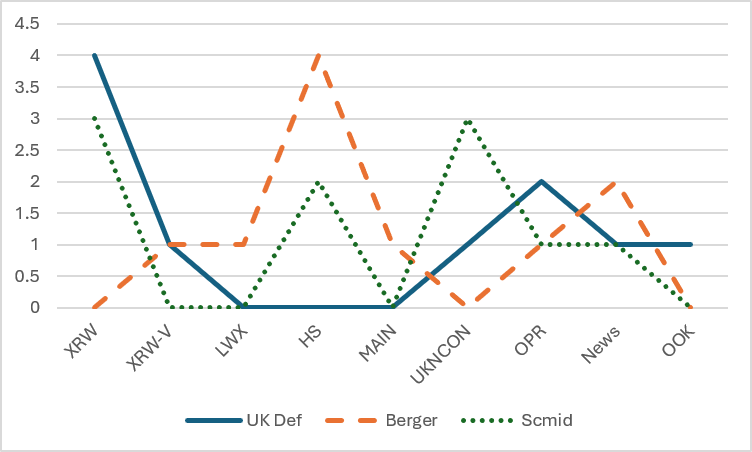}
    \caption{Code comparison by definitional model.
    }
    \label{fig:code-by-model}
\end{figure}

Whilst the coding of the full sample provides a good overview of how all the models work, looking at the definitional models individually provides insight into how each definition impacts classification and filtering. Furthermore, comparing the three models provides a view into how different definitions may be used to produce either broad or narrow results depending on preference and purpose. The coding of the texts broken down by definition is shown in Table~\ref{tab:codes-in-sample}. A~comparative chart of the definitional coding is shown in Figure~\ref{fig:code-by-model}.

\begin{table}[t]
\centering
\begin{tabular}{@{}llll@{}}
\toprule
Text \# & UK & Berger & Schmid \\ \midrule
1 & XRW & HS & XRW \\
2 & OPR & HS & XRW \\
3 & XRW & MAIN & News \\
4 & News & HS & XRW \\
5 & XRW & LWX & UKNCON \\
6 & OOK & News & OPR \\
7 & OPR & News & UKNCON \\
8 & XRW & XRW-V & UKNCON \\
9 & UKNCON & OPR & HS \\
10 & XRW-V & HS & HS \\ \bottomrule
\end{tabular}
\caption{Coding of sample text by definition type.
}
\label{tab:codes-in-sample}
\end{table}

\subsubsection{The UK definition}

The model using the UK Government definition to classify and filter texts provided a sensible range of codes with a relatively good hit rate for extremist texts, with the majority of non-extremist texts falling in line with the definition's prioritization of rights-based, anti-democratic concerns. The sampled texts were dominated by right-wing extremist content including four general texts and one that called openly for violence.

Along with this, texts that were problematic but not extremist were also included, which comprised highly polarised political and generally violent narratives. One text was a news article about a violent extremist group, and one text comprised pushback to extremism and political polarization. In these two cases, it is perhaps the relationality to extremist content in the news story and framing of the pushback that caused their identification. Lastly, there was one text from a non-Anglo/Euro/American context that might be a local or regional form of extremist speech, but without more context this was indeterminate. Interestingly, this was the only model that did not return texts identified as strictly hate speech.

Overall, the classifications align with the priorities of the UK Government definition, though in half the sampled texts this returned non-extremist content. However, three of the five non-extremist texts had other potential harms (political polarization and violence) or were indeterminate, which highlights a needed discussion about both borderline content management and what types of content are acceptable within LLM training data more broadly. 

\subsubsection{The Berger definition}

The Berger definitional model returned the broadest range of texts in terms of coding, with relatively few identified extremist texts. This is likely due to  Berger’s focus on in- and out-group antagonism. These sample texts were dominated by hate-speech-based content, including four texts that included racist, anti-Islamic, and/or anti-Semitic speech.

Interestingly, this sample is the only one that included texts identified as both right-wing and left-wing extremist content. Of these two extremist texts, the right-wing extremist text was identified as calling for violence, and the left-wing extremist text comprised a long and detailed radicalization narrative.

This definitional model was also the only one to classify a text comprised of mainstreaming right-wing extremist narratives, again pointing to borderline content as an issue to be addressed. Two news stories related to extremist events were classified along with one problematic text that although not explicitly extremist is suggestive of violent political sentiment.

Overall, the classifications align with the priorities of the Berger definition, though in more than half the sampled texts this returned non-extremist content. Instead, the texts highlight how definitional priorities impact textual classification. Moreover, this sample suggests a pressing need to explore and address the relationship between hate speech and extreme political positioning in digital contexts. Ultimately, this highlights the need for broader discussions of what types of content are acceptable for inclusion in training data. 

\subsubsection{The Schmid definition}

The texts classified and filtered under the Schmid definition were the most narrowly focused on extremist-identified content and hate speech. This sample also included a number of right-wing extremist texts. However, unlike the other samples, it included multiple international local/regional political texts that were coded as indeterminate without more context (UKNCON). Should these three indeterminate texts be identified as extremist speech, it likely indicates that the Schmid definition is the most useful for identifying non-Anglo/Euro/US texts that are potentially extremist.

Of the remaining texts, two texts were identified as hate speech, one as a news story about terrorism, one as a non-extremist but problematic text related to extremist events, and one as a problematic text focused on xenophobic narratives. Interestingly, the xenophobic text might be considered extremist depending on the context within which the text was originally situated, but given the extractive processes used in making corpora, this cannot be determined further highlighting discussions of borderline content as an issue with LLM training data.

Overall, the classifications align with the priorities of the Schmid definition, particularly given its laundry-list of negative social traits and practices. If the three indeterminate cases can be identified as extremist within their local and regional contexts, this may indicate that the Schmid model is the best at finding explicitly extremist texts. As the coding stands, however, these texts seem related to political polarization, again highlighting the need for broader discussions of what types of content are acceptable for inclusion in training data.

\subsubsection{Summary}

In summary, each model returned texts that matched with its definitional priorities. At best, only about half of the texts returned by the models were identified as extremist content under manual close reading. These various models make clear the ways that definitions impact which texts in a corpus are classified and filtered with variable effects on outputs.

Of note is the complete lack of Jihadist extremist, or even extreme-adjacent texts, with only a single news story about a Jihadist terrorist attack included. This may indicate that stricter regulatory compliance around other extremist ideologies can have a positive effect on publicly accessible data.

Also of note is the large number of texts identified as including borderline content, highlighting the need for more exploration and review of what types of content are deemed appropriate for inclusion in training data. This is crucially important given that the creation of such corpora dislocates texts from their original contexts, which may in some cases obscure the texts' links to extremist speech. Furthermore, because intertextuality can shift the meaning of texts, the reproduction of problematic, borderline content in LLM responses may produce extremist generated outputs, even if explicitly extremist training texts are well filtered. 

These findings also indicate the importance of broader discussions about what is acceptable for inclusion in LLM training data. Whilst explicitly extremist speech may be agreed for exclusion, the variety of sample texts along definitional lines suggests that determining explicit extremist speech is not simple nor necessarily consistently reproducible.

This means that beyond the technical questions around the success of the model, additional questions related to what is explicitly extremist must address how this is determined, such as by group designation status, ideological identification, or based on regulatory compliance (the UK Online Safety Act, the EU Digital Services Act, etc.) requirements. Narrower approaches, even with clearer definitions of extremism, will likely still lead to highly variable model outputs. On the other hand, a broader approach focused on identifying problematic texts (e.g., hate speech, violence, political polarization, etc.) would likely need to address debates over freedom of expression/free speech.
In this case, the purpose of the technologies themselves likely indicates that freedom of expression/free speech are not primary issues as LLM training data is not the same as direct posting by an individual.

Given the existing issues around the efficacy of automated content moderation \citep{Gorwa2020,keydar2026bending}
paired with known issues of bias in training data, this type of content poses similar problems and creates a potential for socio-political harms with respect to outputs. 

\section{Source-domain check}
\label{sec:source_domains}

In order to trace the provenance of the identified extremist documents and check if this information can be useful for screening harmful inputs, we extracted the domains of the documents coded as XR/LW(-V) and HS from the Dolma meta-data (all documents were from different domains). We then computed the number of documents from each of the domains in the whole 200k sample, together with a separate spot-check of the number of documents from the notorious KiwiFarms discussion board, which we knew to be present in the data.

The results, fully reported in Table~\ref{tab:app-domains} in the Appendix, which contains the domain names together with a short characteristic of the resource and document count, show that meta-data statistics have only limited value for pre-screening harmful documents: most evidently extremist resources provided only one document each, with more documents hailing from comments left on different message boards and informational resources. However, the fact that KiwiFarms provided 5 documents to the analysed sample shows that at least some domain filtering was perhaps in order.



\section{Conclusion}
\label{sec:conclusion}

Our aim in this study has been to propose and validate a semi-automated pipeline for identifying extremist texts in LLM training data as well as to provide an overview of such texts, in order to spur further interest in pre-training-data analysis and curation. We believe that our extraction procedure is rather conservative, which, projecting backwards from the results of the final-stage manual analysis, means that roughly half of the documents extracted using the first two stages are likely to contain extremist narratives or hate speech. This puts the estimated prevalence, again conservatively, at $\approx$100 documents in the random sample of 200k documents, i.e.\ the rate of 1/2000. Further analyses are needed to validate the sampling uncertainty of this estimate, but projecting back to the total size of the non-code parts of the Dolma dataset, approx.\, 5~billion documents,\footnote{\href{https://allenai.org/blog/dolma-3-trillion-tokens-open-llm-corpus-9a0ff4b8da64}{Ai2 Dolma: 3 trillion token open corpus for language model pretraining}.} and even allowing for an order-of-magnitude overcounting error, we are still looking at hundreds of thousands of potentially extremely harmful documents in the pre-training pool. Mounting evidence of the limitations of safety-oriented post-training \citep{DahlgrenLindstrm2025,parmar2025} suggests that tackling this challenge at earlier stages of training is a matter of some urgency.

This urgency does not simply relate to ensuring compliant functioning of AI systems, but also to the potential material impacts and harms for individual users and society of not effectively addressing extremist and hate speech content within training data. Concerning instances of self-harm and public violence by AI chatbot users have already been identified in news reports \citep{hill2026}.
Moreover, research aimed at countering hate and violence facilitated by AI chatbots shows that these technologies do assist users in planning violence \citep{killerapps}.
Along with this, research conducted by scholars of terrorism and extremism shows that users are jailbreaking chatbot models and sharing their code on the dark web as a framework for bypassing built-in preventative tooling \citep{molas2024}.

Given this, the findings of the in-depth analysis further indicate an urgency with respect to multi-stakeholder discussions about whether problematic texts, particularly violent and politically polarised texts, as well as extremist and hate speech texts are appropritate training data for LLM-based interactive technologies. This work provides a foundation for those initial discussions. While such discussions are ongoing, continuing development of technical responses is essential.  

On the technical side, one interesting avenue for future work is reverse-engineering the \enquote{operational} definitions of extremism used by the models in practice. As noted in Section~\ref{sec:prevalence}, Qwen~2.5 seems to return a lot of false positives while being very good at applying the union definition as the actual union of individual definitions. This indicates that it has a robust but opaque set of criteria for classifying extremist texts of different types, which somehow do not perfectly align with the letter of the provided definitions.


\section*{Limitations}

The main limitations of the study are twofold.

Firstly, due to difficulties with rechecking documents rejected by the first stage of the analytic pipeline, our method is likely to produce false negatives. Therefore we can only provide a lower-bound estimate on the prevalence of extremist documents in the training data.

Secondly, due to low prevalence rate of extremist documents in the analysed sample (1/2000), our estimate is likely to be affected by high sampling variance. However, given that our initial document sample is i.i.d.\ and that our results are significant \textit{even if we overcount extremist documents by a factor of ten}, this limitation does not seem crucial at this stage of analysis.



\bibliography{custom}

\clearpage
\onecolumn

\appendix
\renewcommand{\thesection}{Appendix \Alph{section}}

\section{Prompts}
\label{sec:appendix-prompts}

\subsection{By-definition prompts}

As embedded in Python code:

\begin{lstlisting}[language=Python,breaklines=true]
EXTREMISM_DEFINITIONS = {
    "uk": """Extremism is the promotion or advancement of an ideology based on violence, hatred or intolerance, that aims to:
1. negate or destroy the fundamental rights and freedoms of others; or
2. undermine, overturn or replace the system of liberal democracy and democratic rights; or
3. intentionally create a permissive environment for others to achieve the results in (1) or (2).""",
    "berger": """Extremism refers to the belief that the success or survival of the speaker's or author's in-group can never be separated from the need for hostile action against an out-group. The hostile action must be part of the in-group's definition of success. Hostile acts can range from verbal attacks and diminishment to discriminatory behavior, violence, and even genocide.""",
    "schmid": """In the context of democratic societies, extremism is a form of political expression (usually on the far left or the far right of the political spectrum) that is not acceptable to the more moderate mainstream of political life. Extremist groups and parties tend to be anti-constitutional, antidemocratic, anti-pluralistic, fanatical, intolerant, non-compromising, single-minded, authoritarian and adhering to an ends-justify-means philosophy, wanting to realise their goals by any means, including the use of political violence against opponents. Extremists on the political left or right and those of a religious-fundamentalist orientation favour violence over persuasion, uniformity over diversity, unity over pluralism and orders over dialogue.""",
}

SYSTEM_PROMPT_TEMPLATE = """You are an expert in political language. Check if the document below satisfies the following definition of extremist speech:

{extremism_defn}

Use only the content of the document. Ignore source or author.
Respond with one word: "Yes" or "No".
"""
\end{lstlisting}

\subsection{Union prompt}

\begin{lstlisting}[breaklines=True]
You are an expert in political language. Check if the document below satisfies any of the following definitions of extremist speech:


Definition 1:
Extremism is the promotion or advancement of an ideology based on violence, hatred or intolerance, that aims to:
1. negate or destroy the fundamental rights and freedoms of others; or
2. undermine, overturn or replace the system of liberal democracy and democratic rights; or
3. intentionally create a permissive environment for others to achieve the results in (1) or (2).

Definition 2:
Extremism refers to the belief that the success or survival of the speaker's or author's in-group can never be separated from the need for hostile action against an out-group. The hostile action must be part of the in-group's definition of success. Hostile acts can range from verbal attacks and diminishment to discriminatory behavior, violence, and even genocide.

Definition 3:
In the context of democratic societies, extremism is a form of political expression (usually on the far left or the far right of the political spectrum) that is not acceptable to the more moderate mainstream of political life. Extremist groups and parties tend to be anti-constitutional, antidemocratic, anti-pluralistic, fanatical, intolerant, non-compromising, single-minded, authoritarian and adhering to an ends-justify-means philosophy, wanting to realise their goals by any means, including the use of political violence against opponents. Extremists on the political left or right and those of a religious-fundamentalist orientation favour violence over persuasion, uniformity over diversity, unity over pluralism and orders over dialogue.


Use only the content of the document. Ignore source or author.
Respond with one word: "Yes" or "No".    
\end{lstlisting}

\section{Source-domain statistics for extremist/hate-speech documents}
\label{app:ssec-domains}

Table~\ref{tab:app-domains} shows the prevalence in the analysed 200k document sample of the domains that provided the documents identified through close reading as containing extremist narratives or hate speech.

\begin{table}
\begin{tabular}{p{5.2cm}rp{8.5cm}}
\toprule
\textbf{Domain} & \textbf{Count} & \textbf{Description} \\
\midrule
gloria.tv & 3 & Traditionalist Catholic media platform with conservative religious commentary. \\
teapartyeconomist.com & 2 & Tea Party and libertarian-conservative political blog. \\
discordleaks.unicornriot.ninja & 2 & Investigative archive of leaked Discord chats from extremist communities. \\
alexjones101.blogspot.com & 1 & Personal conspiracy-oriented political blog. \\
meforum.org & 2 & Middle East Forum, a neoconservative think tank focused on Middle East policy and Islamism. \\
us5.campaign-archive.com & 1 & Mailchimp-hosted archive of email newsletters. \\
ilanamercer.com & 1 & Personal blog of paleoconservative/libertarian commentator Ilana Mercer. \\
chechar.wordpress.com & 1 & Extreme white nationalist blog. \\
leozagami.com & 1 & Personal blog promoting conspiracy theories and esoteric interpretations of current affairs. \\
debka.com & 4 & Israeli news and geopolitical analysis site emphasising security and intelligence topics. \\
cunningrealist.blogspot.com & 1 & Independent foreign policy and political commentary blog. \\
en.uncyclopedia.co & 2 & Satirical encyclopedia parody website. \\
chuckontherightside.blogspot.com & 1 & Conservative political commentary blog. \\
therabbith0le.com & 1 & White identity/white nationalist website focused on \enquote{white victimhood} narratives. \\
liberation.ndfp.org & 1 & Official news and commentary site of the National Democratic Front of the Philippines. \\
\midrule
kiwifarms.net/kiwifarms.st & 5 & Internet forum known for coordinated harassment campaigns. \\
\bottomrule
\end{tabular}
\caption{Prevalence of source domains that provided manually-verified extremist and hate-speech documents in the analysed Dolma sample + a spot-check for KiwiFarms.}
\label{tab:app-domains}
\end{table}

\section{Manually analysed documents}
\label{sec:appendix-sample}

\subsection{The UK definition}

\begin{enumerate}
\item Michelle Obama will travel to South Africa later this month. The First Lady’s trip coincides with the release of my new book, “Into the Cannibal’s Pot: Lessons For America From Post-Apartheid South Africa.” And not a moment too soon. (Read the Preface on VDARE.COM.) “Into The Cannibal’s Pot” will dispel any myths Michelle Obama is likely to help perpetuate about this writer’s former homeland. So why is this book so very crucial at this juncture in our history? Simply this: It is essential that we curb the naïve enthusiasm among American elites, and those they’ve gulled, for radical, imposed, top-down transformations of relatively stable, if imperfect, societies, including their own. As the example of South Africa demonstrates, a highly developed Western society can be dismantled with relative ease. In South Africa, this deconstruction has come about in the wake of an almost overnight shift in the majority/minority power structure. In the U.S., a slower, more incremental, but equally detrimental, transformation is underway. Americans, moreover, should know what I divulge in Chapter 7 (“The Anglo-American-Australian Axis Of Evil”): Washington and Westminster bear considerable responsibility for the “swelling social disorder” in South Africa, having insisted that South Africa pass into the hands of a voracious majority. Unwise South African leaders acquiesced. Federalism was discounted. Minority rights for the Afrikaner, Anglo and Zulu were dismissed.Ironically, America’s founding fathers had attempted to forestall pure democracy by devising a republic. Yet under the wing of the American eagle a dispensation was negotiated in South Africa, the consequence of which is the raw, ripe rule of the mob and its dominant, anointed party. The time is thus historically ripe to challenge some of the central tenets of a liberal democratic ideology that would bring about the disaster that is post-Apartheid South Africa.Incredibly, the country was scorned by the West and treated as Saddam Hussein was, with boycotts and sanctions when it was governed by a racist white minority. Now that a racist, black-majority government controls the country; that it is as violent as Iraq, Liberia, or the Congo and rapidly becoming another Islamist-friendly, failed African state, it is the toast of the West. Indeed, world leaders and the liberal lickspittle media seldom speak of the embarrassment that is the democratic South Africa—the crumbling infrastructure of this once First-World country, and the out-of-control crime. Rocker Bono certainly isn’t moved to tears over the (seemingly) systematic extermination of the Afrikaner farmers of South Africa. The cultural cognoscenti in the US are equally silent about the New South Africa’s unparalleled, radical, race-based wealth-distribution policies. More people are murdered in one week under African rule than died under the detention of the Afrikaner government over the course of roughly four decades. Americans, who take for granted their domestic tranquility, can’t afford to finesse the fate of the dying Christian civilization at the tip of Africa. “Into The Cannibal’s Pot” compels them to stare into “The Heart of Darkness” that is the New South Africa, and by so doing, offers a cautionary tale:In their unqualified paeans to the will of the majority everywhere, Americans must understand that universal suffrage is not to be conflated with freedom. As the democratic South Africa (and Iraq) amply demonstrates, political rights don’t secure the natural rights to life, liberty, property, and the pursuit of happiness; ink-stained fingers don’t inoculate against blood stains. Extant societal structures that safeguard life and property can always be improved upon. But once these bulwarks against mob rule and mayhem disintegrate, they are seldom restored. A civilized society, ultimately, is one in which the individual can go about the business of life unmolested. If he can’t do that simple thing, of what value is the vote? The Apartheid-era, traditionally Western legal institutions, however flawed, were preferable to the Rambo Nation’s “rehabilitated” institutions, riven as they are by tribal feuds, fetishes, and factional loyalties. America’s intellectual “Idiocracy”—the president and the “Untamed Ids” of the media, liberal, libertarian, and conservative—are egging on revolution in the Middle East. Post-apartheid South Africa should serve to remind this retinue of romantics that stable societies, however imperfect, are fragile. They can, and will, crumble in culturally inhospitable climes. For better or for worse, societies are built slowly from the soil up, not from the sky down. And by people, not by political decree. Our unhappy trek through the wreck of the New South Africa begins with the facts, nothing but the facts. The realities of crime-riddled democratic South Africa are relayed in Chapter 1: “Crime, the Beloved Country.” The title parodies Alan Paton’s poignant tale titled “Cry, the Beloved Country.” That story was to apartheid South Africa what Harriet Beecher Stowe’s “Uncle Tom’s Cabin” was to antebellum America. Chapter 2, “The Kulaks of South Africa Vs. The Xhosa Nostra,” provides a brief, action-packed, history of Boer, Briton and Bantu, before moving on to the ethnocide the FLOTUS won’t mention: Afrikaner farmers are being culled like springbok in a hunting safari. South Africa is a microcosm of what America could become, unless it returns to the principles that made it great. If American institutions continue to subordinate their raison d’être to politically dictated egalitarianism, reclaiming them from the deforming clutches of state-enforced tribalism will become harder and harder.
\item You were strong, trained, grown men who signed up. You were scared of disobeying orders, and a little boy with a pea shooter. You deserve this. Cowards who disobey their oath to protect others to blindly follow orders while the children of their community get killed, they’re actually getting off easy with just receiving threats! Post all the officers contact information and addresses. They need to be held accountable instead of getting a slap on the wrist. I would never threaten them personally, but do I feel like uniformed police who failed to act and let 19 people mostly kids die…. deserve to die themselves? Yes. They deserve nothing less than the death penalty for the level of cowardice shown to their community. In reality stop voting blue or senators and elected who don't push for ope records and ipra like new mexico or other sunshine states. Bullsh1t. Most wear radios without clip. It goes in holster. Did job 18 years. Bad. What have they learned? I want to send you a patch from my home town. I’m not a huge fan of Bill Maher, but he said it best: We give cops all kinds of power and authority because they supposedly are risking their lives for our safety. Well, in exchange for all that power, we expect them to step up on the rare – or, sadly, not so rare – occasion that sh*t hits the fan. When they fail to do so, it makes people wonder what cops are really for. Small, low budgeted and inexperienced police departments is what Uvalde breaks down to. Chiefs of Police for places like schools districts are usually retired officers from somewhere else, or left a department late in their career for more money (which I doubt). I don't know that the information is out there on his employment history but whether or not he was competent is one thing, but he did, in my opinion act recklessly and irresponsibly. Any police officer worth his salt will not try to take command of a scene when he is not even fully equipped or able to communicate or at the very least knows everything that the officers on the inside know. Who knows how long it was from the time the shooting began to the time the Chief first began communicating and receiving information about this incident. I was not there on the inside when this was going on so I can't speak to it with any factual backing, but I can speculate based on what I know as of current. Since Columbine, policy has changed regarding active shooters/school shooters. We had general standing orders that if responding to an active shooter situation, priority one is stopping the threat by any means necessary. This includes going in alone and without backup if the circumstances are obviously exigent and that order needs to be followed.  I just know that with the agency I spent my career with, we would have not waited for an incident commander, but took action under general orders, or from one of our immediate superiors (normally a Sergeant, Corporal, or OIC with a significant amount of experience). I don't know what happened in there, but I know that putting your life on the line in defense of yourself or others is something we all knew when we signed up. Good episode until you said you can't shout fire in a theatre. You absolutely can and prosecution for it would absolutely be thrown out before trial. In hick towns like this, most officers dont have lapel mics. The dept is too cheap to provide them. Hell even LAPD makes them use hand helds. Insane.  I'm from Albuquerque, born and raised!!! You totally glossed over the drowning story in addition to making excuses for those tyrants on scene of that incident. body cam footage is available for viewing, , And the investigation they were conducting exceeded any level of reasonableness once they determine that no crime had been committed essentially harassed that man, Who in addition to having emotional issues was reportedly also homeless. This is what police all over this country do to American citizens in this type of situation. When you watch ALL of the footage you will see these officers do nothing to stop him from getting into the water before he got into trouble. Even more disturbing is the vantage point of one of the body cams that show The son of a bitch cop resting his arms on the rail, casually asking how far out do you think he’s going to get?  And the cops that failed the children. They ought not ever to be eligible for any position in law enforcement throughout the country. Blackballed not only from law enforcement but any position as a public servant. There is no greater proof that those in law enforcement failed the children and failed those who pay their salaries with a false sense of security. Do not allow these parasites of the system to indulge themselves under false pretenses in the coffers we fill with our hard-earned money. If there was ever a time to make a point this is it. Every public employee must recognize who they work for before they are hired. These pompous individuals have the empty-headed belief they are doing us a favor when in fact it's the exact opposite! Maybe we spend money on mental health treatments. It's apparent this guy needed psychiatric help. The guy jumped into the water of his own free will. He drowned and that 100\% on him. The cops have zero accountability for this. I'm a disabled Firefighter/EMT/911 an officer dispatcher. Cops need to be trained for self defense before even thinking about doing my old job. If the cops would have jumped in to save him without the proper training and equipment the chances are he would have taken one of them or both down with him. It's just not worth it. Sad but that's reality. Second amendment is civil right..NOT UP FOR A VOTE. That is what the ACLU and the gang have created. INCOPRORATION OF THE CONSTITUTION on the state…so now NO state can vote away what ONLY an amendment can. Remember 6ay so called marriage…IT Nullified 40 state sprohibition against 6ay so called marriage…the federal courts ERASED that right. So now the second amendment is just as sacrosanct..THANK YO, ACLU. Most cops are cowards that no NOTHING of the Constitution and Bill of Rights and could care less to learn even though they swear an oath to uphold. If someone jumps off a cliff should the cop also jump and try to catch them. You will not see body cam footage because it was faked. Quick question: you guys had a video titled: fake cop causes extreme road rage incident (James Hoefert) – from when this channel was called fire and police videos, why is it private now? 1:40 of your intro banner? Unsubscribed. Tiny town with a 'can't happen here' mentality. There are thousands of places like that. Complete FUBAR of base protection and response. This situation was lost when the school was not secure. So you think patriotic right leaning American citizens are all inbred? Because that is what you said. Maybe it was word salad and not what you meant, but uhm yeah. Conspiracy to riot…. Why hasn't Maxine Waters or Kamala Harris been arrested? I'm not sure I get you. You seem to lean left but are a blue line supporter, or appear to be. You make a lot of excuses for bad policing, when it infringes on citizens rights. The reason we have lost a lot of our rights is by tiny increments, of people saying "well it does violate this person's rights, but we really need to get this bad guy in a cage. So just this once we will let it slide" Boom precedent is set and now everybody is subject to that little bit of lost rights. I watched you before I subscribed, lots of content, and I still don't get you. People except so much from cops and yet are attacked on all fronts constantly. They put themselves in harm's way every shift for very average remuneration. Thanks to all the heroes! Dereliction of duty. Bunch of cowards, the lot of them. Unlike paramedics and EMT's, cops have no legal duty to act (in the case of law enforcement, protect) while on the clock. Cops do NOT exist to protect you. That is not their job. Cops are the "force" arm of the .gov. They are only there to do .gov bidding.
\item 9/11 (US), 7/7 (London), 3/11 (Madrid) and 11/26 (Mumbai). Perhaps your action slow, and eventually stop, the site meter of Islamic terror attacks, which stands at 12690 today since 9/11. This meter is running very fast. We do not know who will fall victim to the next Islamic terror or suicide attack: maybe me, maybe you. Your sharing this e-book with your friends, colleagues and contacts may reduce the chances of you and I be the next victim. We at the Islam watch and faith freedom are working hard to save the world from the atrocious demon, called Islam. The work done here is voluntary. Dr. Ali Sina and M.A. Khan have left their jobs and working on this most pressing issue of our time at great personal sacrifice. You can support and help this venture in any way possible: you can promote us, give our link to others, talk about us or share our contents in other forums. I was born in Pakistan as an atheist. Then Islam, a cult, was imposed upon me as a religion. Muslim invaders who conquered large parts of my country forced my ancestors to convert to their religion. Once one generation succumbed to Islam, the successive generations were either threatened or brainwashed so that they would never leave this evil cult. My forefathers were even used by Muslim invaders to spread their evil faith and empire. I was a fundamentalist Muslim with beard. I, however, managed to realize that Islam, in the guise of a religion, is actually an imperialistic ideology, similar to Nazism. It is like a net with the spider waiting for the big meal. It spreads through aggression and deception. It is time that we all try to understand 'what Islam truly is' and unite humankind to confront this vicious doctrine of hate. This book is a small step in that direction. Pertinacious ignorance and cover up of evil in islam isn't worth perdtion of one soul. MOCNE ! Ks. prof. Guz - O operacji ,,ZARAZA19" - jak ratować swoją duszę ? Komu pierdziel, komu dzwon. Niszczycielom Katolickich Świąt Wielkanocnych. Dziewięć osób z domu opieki koło Puław zmarło po zaszczepieniu. Czy śledztwo trwa?
\item Mazar-I-Sharif, Afghanistan - Afghan clerics and their followers threatened violence against the government on Sunday over the release of a Christian convert, saying he had to be brought back from Italy and put on trial. There has been fiery criticism of the government over the release of the convert, who was spirited out of the country last week, but protests have been few and peaceful. The convert, Abdur Rahman, 40, was jailed last month for converting to Christianity and could have faced trial under Islamic sharia law that stipulates death as punishment for apostasy. After a storm of Western criticism, led by the United States, Rahman was released and taken to Italy. About 1,000 people gathered in a mosque in the northeastern town of Kunduz and demanded that Rahman be brought back from Italy and sentenced to death. "This act of the government is illegal," Sheikh Mohammad Baqir, a cleric and organizer of the rally, said, referring to Rahman's release. "Either he should be tried or the government should go. We urge other provinces to raise their voices and if the government doesn't listen, we will resort to violence," he said attracting calls of "Allahu akbar" (God is Greatest) from the crowd. Police refused to let the gathering leave the mosque and march through the town. A police official said they were worried about violence if a march was allowed. Afghanistan saw violent protest in February over cartoons of Islam's Prophet Mohammad published in European newspapers. Violence also broke out last year during protests over a magazine report U.S. military interrogators had desecrated the Koran. Many conservatives in Afghanistan had insisted Rahman be tried under Islamic law. The lower house of the Afghan parliament also said his release was illegal.
\item Always had to affirm his status with everyone. Hell even sieg heiled random people for kicks because he could get away with it. Uppity rich niggers are worse than Hoodniggers in my opinion. high trust society is synonymous with successful society. So pretty much all societies worth infiltrating are able to be infiltrated.  @Mr. Wright that is interesting how quickly they become disinterested in social justice once they get rich. Because they have it all. Duh. But as I said before they are *instinctively* insecure even if they dont realize it.  "I had but inbested in bitcoin" Actually he was a wingnut due to testes cancer.  Happanigger I think. Actually comparatively decent looking. You know what I think I saw today? I didn't look too close. I think I saw a South East Asian granny and an Indian granny with their mixed race grandchildren. Bangladesh and Laos namely. Once I became fully Race Realist alot of the guy's tendencies made sense. Also he was Ironically our groups token black guy in every sense of the word. Reason why I mentioned it so much is that its an interesting Empirical study. Hopefully you're right. I saw him ask his followers to report you. Bantzing talib is always worth it. JDF is my favorite news site. Am I the only one who legitimately had no clue about this before today's Merchant Minute? @Hector Iron March dudes warned about this months ago. We need to just let Venezuala burn. Don't take any refugees. Make them go to colombia or brazil. yeah, taking in refugees is the most retardedly disgenic practice. "oh your whole society failed \& you are the most fail members of that society? come right on in!" I mean he belives Fox News is a white supremacist site too. @Zorost that myth about black women being the most educated group... yeah, its ridiculous. Nuance is a white social construct. Is this *the* lauritz? you buying the dip!?!?!? Hey what is that think tank Mike Enoch is involved in? @Lauritzvongh so it doesnt have a minimum buy in? doesn't that depend on whether you use a bondsman rather than the court? @NSJW If anyone knows the name I think I need to get in touch with them asap. I may have something they will want to be aware of. @Zorost like every weekend in the hood?? wasn't he never actually convicted of cannabalism?
\item By all means, keep telling yourself that fantasy right up until this nation's people rise up and impoverish you.  Russian and French elitists once deluded themselves with this egotistically driven fantasy too and lost their fortunes/lives over it.  World history teaches that economic and political disenfranchisement always ends poorly for those who make it possible.  Current trends in the U.S. will end no differently.    Make no mistake, this nation is a powder keg about to go off due to the political and economic disenfranchisement plutocrats have seen fit to increasingly institute over the past 30 years.  It will only take a minor spark to set it off and when it blows, there will be precious few plutocrats and minions who will survive it untouched, regardless of the "safe harbors" they mistakenly believe will protect them.  The last time right wing ideologues destabilized the global community, as you favor doing, it culminated in either major revolutions and/or a World War.  EVERYBODY loses in that scenario.  One last point, communism isn't gone.    The suggestion that there's a lack of options is as short-sighted as one's social views can get.  The American people have a LOT of choices at their disposal and if ideologically-driven, narcissistic fools persist in denying them those choices at the ballot box...a bloody revolution will be inevitable.  It always ends that way...  Since it's lost on you, the American people are more than happy to make billionaires extinct since their presence threatens the American Dream.
\item Local serial rapist and child molester Brent Brents has been busy writing letters from prison. Can we please throw this piece of shit into a holding cell with some hardcore gang-bangers from Aurora and then look the other way when they sodomize and jab a sharpened spoon stuck into his neck? 3 Comments I second that notion, matthew. I am watching a movie before bed lately where a guy gets a spoon plunged into his neck. It doesn’t seem that hard to do. I say we televise on Pay Per View. You have a dual advantage: pay off the court expenses and investigative work, and deter future child molesters … hopefully.
\item Even if the Norway terrorist was REAL the media and social engineers will manipulate it to serve their interests. Point and case. They altered his facebook from being a mason with some liberal views all of a sudden to a crazy right wing extremist. Like Alex quoted a politician. "Can't let a good tragedy go to waste" I kinda sided with Dr Pitchenik today. I think the norway terrorist was a true terrorist.Its time for AMERICANS that are AWAKE to RALLY together and get ready for a new kind of REVOLUTION because these people in HIGH places are not going change their plans of taking us over and making us slaves in the NEW WORLD ORDER. We need to send out a MAJOR message that THEY MIGHT HAVE THE POWER IN CONGRESS, WALL STREET, TREASURY, NATO..etc BUT WE HAVE THE POWER HERE ON THE LAND!!! ITS ABOUT TIME TO OVERTHROW THIS GOVERNMENT!!!! and their SUPER RICH ALLIES
\item Kanthar Balanathan, Australia DipEE(UK), GradCert(Rel Eng-Monash), DipBus\&Adm(Finance-Massey), CEng. MIEE. Tamil Nadu (TN) changed its name from Madras state to Tamil Nadu when CN Annathurai (CNA) of DMK assumed power. The only non-Dravidian party that held power was Congress and the Chief Minister was the then Kamarajar. The first post-independence assembly power was held by Indian National Congress (INC) by C. Rajagopalachari and Kamarajar, and the INC held power until 1967 when CN Annathurai of Dravida Munnetra Kazhagam became the Chief Minister. Neither Congress or BJP is unable to capture power in the State of TN to-date. Tamil Nadu state consists of several millions of Tamils and is a historic state where India cannot ignore their demands. Tamils have participated in the freedom struggle and Kamarajar was an influential politician in the INC. With the creation of the Dravida Kazhagam (DK), Dravida Munnetra Kazhagam (DMK), and Anna Dravida Munnetra Kazhagam (ADMK), DMK \& ADMK was popular among the Tamils of TN. From 1967 to-date, either INC or BJP is unable to capture the power of the state government. India made several silly political blunders where they misinterpreted the political strategy and the capability of the Ceylon government in the 70s. Ceylon planned to declare as a republican state and it is assumed a foreign government, (UK) planned the JVP insurrection in 1971 which was demolished by the then late PM Srimavo B. However, the Ceylon government knew that they have another front of terrorism imminent from the North. In 1972, the Tamil New Tigers (TNT) was declared by the late Vellupillai Prabakaran. GOSL had several fronts of terrorism emanating in multi-dimensional directions. This was not due to people’s desire, however, as a result of foreign involvement to disturb the peace and stability in SL. The big question is: Did the UNP help the two terrorist fronts to disturb the peace so that they can crawl to capture power? We know that the JVP and LTTE worked together as concealed fronts for the attacks. It cannot be disputed that wherever the UK had its colony and gave independence, it left a major dormant problem knowing that the UK can hold onto the economic domination of these countries. Some countries gave into the empire’s commanding \& bullying, However, SL with Srimavo and the SLFP ideologies opposed to the domination in the 70s. The Biafran war was a result of the British involvement in Nigeria where hundreds of thousands of Ibos were massacred by the military. Most of the JVP clowns ran to the UK on defeat. Subsequently, the LTTE gave birth in 1976 by changing the name from TNT to LTTE. Prabakaran was NOT an intelligent person, however, violent during his young ages and also with his uncle Gnanamoorthy in the FP, he (VP) was brainwashed.  The association of most youngsters and males with the South Indian smugglers and the stunt movies brainwashed the minds of the youngsters to engage in violence. Tamil Nadu is another racist state, helped the LTTE to expand with funds and weapons. A RAW key personnel met Prabakaran in 1984 after the LTTE attack on the 13 soldiers in Tirunelveli. This meeting took place in Pondicherry. Another question is why was the Tirunelveli attack planned and executed by Prabakaran? Was RAW and the Indian government behind this project to blow-up turmoil, pandemonium in SL? INC politicians tried with maximum planning and merciless activities to capture power in TN state by bringing in the INC government, to state, helping the LTTE through MGR, the movie star who was the CM. This may have been on the instruction from late Indira Gandhi. MGR is said to have given Prabakaran Crores of INRs and allowed to operate freely in TN state for training in Coimbatore and other places and administer the terrorist activities. It was easy for Prabakaran and the LTTE to travel through the gap between Tuticorin and VVT easily. Another blunder Rajiv Gandhi made was to propose the 13th amendment. Another mistake was JR was about to grant the 13th amendment fully. The author of this article lived in VVT and his father was the OIC/VVT of the Police force in 1958 for approximately ten months. Another political error Srimavo made was to be responsible for the transfer on my father form VVT out as a result of a petition by a VVT smuggler. Whatever, INC tried they could not capture power. On the other hand, BJP wanted to capture power in TN and organized various political activities, however, BJP failed. People like Nedumaran, Vaiko, Seeman, etc are boys of the two major parties. The reason for the two major parties to always talk about implementing the 13th amendment fully is to attract the votes in Tamil Nadu in their favor. It is a cunning strategy India adopts to win the support of the people in TN. It is not because of the love towards the Tamils in SL. India wants to dominate SL. TN politicians’ campaign about SL Tamils autonomy is to win elections and be the governing party. India’s INC/BJP campaigning for SL Tamils is to enter into TN and be the governing party. It is certain that neither INC or BJP will never ever enter to govern TN. Further Indian government whether its INC or BJP campaigning for SL Tamils is to threaten SL and prevent SL to have close ties with China. India has a fear of China’s invasion via SL. India does not know that China has already invaded India economically and changing TN culture. If we go back to history, all the politicians mentioned above were arrested and released. On what grounds were they released? It is to fool SL. Recently we understand that Mordi has asked our President HE GR to implement the 13th amendment fully. Modi removed the autonomous powers of Kashmir, however, on what grounds is he asking HE GR? In plain words, Modi has no business to interfere with the political activities of SL. He has to mind his own business. Modi and the former CM of TN, Panneerselvam were Tea stall keepers who crawled into politics: Don’t know how? Can we assess the intelligence and the knowledge these two have to rule a big country like India? They practice cheap politics. AS far as SL is concerned the UK and India are the biggest problems and a threat to SL. Not the Tamils. With all these activities, Delhi is settling millions of Hindi speaking people in TN. Major industries are owned by Hindis. The majority of Tamilians can speak Hindi, Telugu, Malayalam, Karnataka. It is speculated that about a million more Hindi speaking people have settled in TN. Every Tamil movie is contaminated with Hindi. The idiocy of the TN Tamilians is to talk about SL Tamils, however, not about their own state being swallowed by the Hindis. Tamilians in TN are Saivites. What India does not understand is that the West may have already planned to divide India into several republics, and will never allow to grow like China and become a threat to the rest of the world. Seeman is associated with the movie world acting in movies and making movies. Indians are mindset and fanatic in movies and they think what the hero does is real. That’s how late MGR \& Jayalalitha became CM. Both are uneducated. Maybe it’s a common decease. In the USA Arnold S, Clint Eastwood was drawn into politics. People do not realize that acting is a skill and a make-believe imitation. This is one of the major weaknesses in Democracy where numerical superiority overrides capability \& competence. There may be three million Tamils in the UK. Not all three million people like SL divided. There are patriots who love SL as a unitary state. The three million Tamils will not forgo their comfortable living and some with the DOLE they receive return to SL. What business has UK got to include in the Conservative’s manifesto that SL should be divided into two states? Cheap Politics. Is the UK prepared to divide the UK into four republics? Is the USA prepared to divide the USA into several republics? The weakness of the Indian and the UK government is to talk about HR violations, missing persons, division etc. They should mind their own business and make their country stable and develop the countries. UK and India have adequate political social problems in their own countries. The UK may be converted to an Islamic state in a few years. Nehru wanted Ceylonese leaders to realize that Ceylon might not be necessary for India, but India was necessary for Ceylon. Who said that Nehru was a clever politician? Maybe Gandhi was and Professor Subramaniam Swamy is intellectual. There are so many concealed mysterious weaknesses among the Nehru family anyway. THEREFORE, INDIA SHOULD STOP INTERFERING INTO SL AFFAIRS AND NOT BE A SLAVE TO TN IN THE PLOY OF CAPTURING PARTY POWER. SL should examine and revoke the 13th amendment for the following reasons. Granting Police \& Land power will disturb the equilibrium of the governing authority of the parliament. Tamils are not politically and socially matured enough to exercise police and land power. With police power, they want to control their politics. Tamils have differences with respect to caste, religion and race and they are biased. The writer knows the atrocities Jaffna man committed against Indian children in the name of domestic workers. Jaffna man does not like or want to associate with Indian workers from the highlands. Tamils are always on the lookout to run away from SL. What advantage does Tamil Diaspora have in getting Federalism or Tamil Eelam? Even in foreign countries Tamils are not united but exercise caste differential. Giving Police and Land power will lead to another form of internal terrorism and destroy the united SL, as Tamils are economically a volatile race. What Tamils need in SL is maybe a small-scale autonomy without police and land power. Already the Provincial Councils have blown up the expenditure and driven SL towards bankruptcy. Arjuna Mahendran is a typical example. With the already granted NPC, what has C.V. Vigneswaran done to the North? Only jabbering and blathering about missing persons and Tamil Eelam. Now he is winding up young people to form his own party. Corruption and bribery were maximum in the North. A new committee should be formed to study and formulate a system of how SL could be governed. In mind that India revoked Kashmir’s powers of the forthcoming threat to its stability. SL should never, ever, mind about Indian intervention and UK blabbering. Indians are political fools. If they do not mind their own business then China may teach them a lesson on how to govern and mind their own business. People like Sritharan, Gajendrakumar Ponnambalam, And Gajendran are a pitfall of the SriLankan society. The UK massacred millions of people wherever they ruled. They plundered those countries. Now they talk of democracy, justice and humanism. Shame on them. Tamil Diaspora is eager to rip Tamils to become rich. By talking that Prabakaran is alive they have been plundering wealth from the Tamil FOOLS. Above is an example of a group where some members ripped the association thousands of pounds. This is one example. In Australia, some who associated with LTTE, and other associations, like Seniors have ripped several million. Federalism or a separate state for the Tamils in SL is a risk for the region. If the UK has love towards the Tamils, they should relocate all the SL Tamils and those in the UK and settle them in an island suitable to the crowd, such as Madagascar, or Reunion island or even Diego Garcia. E.g. The Jews were relocated to Israel. Tamils: We are an insane race judging by the standard of CV Vigneswaran, Mavai, Sambanthar, SJVC, etc. GOSL has every right to build Buddhist temples, and settle Sinhalas in the N\&E. Its Democracy.
\item White people: You got to listen to me.  But this is serious. The stinking bastards are now out to destroy our history and rich White heritage. And right before our eyes, too! OK, I get it.  But did you ever expect your own race to get openly written away, slowly perhaps, but surely? They know they can get away with it. They’ve worked to brainwash us with “diversity” BS for decades. Like trained parrots, they can’t talk for a minute on TV without saying “diversity is what makes us so great.” Total BS. America was founded by great White people, Christian Europeans. Yep, Christians. The Founding Fathers were not “pantheistic” but simply trying to keep foreign religious powers out of our America dealy. We can see the DC Zionists in the thrall of Israel have now completely ignored all that. Now I’m just a regular guy. Certainly no deep thinker or great writer. But I can put two-and-two together to see where all this is going.  Us Whites have done great things — our inventions alone are magnificent. Our efforts to help the non-Whites of other countries is readily apparent. Billions are simply alive right now because of all we’ve done. If you think it out just a bit, you too will be totally pissed over the nerve of these dirty backstabbing creeps. CAN YOU BELIEVE THIS TOTAL BULL CACA? Ursula Haverbeck is a 89 year-old woman in Germany who was put in prison this week over what she simply said in public about the holocaust. She had the nerve to say on live German TV she didn’t believe it happened like they tell us. She had sincere doubts about the whole gassing part really happening. This old lady (just look at her above) should be in a nursing home, not a jail. They sentenced her to 2 years WITHOUT PAROLE. The poor woman apparently tried to get away. She was due to report for prison on May 2 but went missing. The dirty Jews screamed for “high pressure” searches for the “Nazi grandma.” Finally, she showed up at her home and was promptly bagged like a dangerous fugitive. Just sit there and stew on all this for a moment.  I already knew about the story and wanted to write about it here.  I get emails about all this from a certain someone who himself has been arrested and attacked mercilessly by filthy Jews in another country. Little White schoolgirl gang-raped by muzzie men in a cemetery in broad daylight. Yesterday, the US Boy Scouts dropped the word ‘boy” from their name. They’ve already allowed chicks in, along with open faggots — including unit leaders they go out camping with in the woods (imagine that). I used to be a Cub Scout (junior level boy scout). When I started going, the mother running our particular cub unit would have us clean her entire home. That happened every single meeting for at least a couple of months. That’s all we ever did. She was clearly using us as her little domestic slaves, so I dropped out. I wanted hiking, camping and shooting real guns. But by then, I was pretty much already a wild child (we lived next to the mountains) and could get my woodland groove on all by myself. My younger brother joined something called “THE INDIAN GUIDES.”  I remember them doing this downhill race where the kids would build a wood car to fit both dad and son — kind of like the Soap Box Derby but with few real construction rules or any course. A disaster waiting to happen. My dad and my brother built a car to race. It was a fairly standard thing — maybe boring but safe. When I went to the race there was this one father who had built his son this cool-looking batmobile thing that everyone was oohing and ahhing over. Even had a nice purple paint job. But I noticed that it didn’t have any brakes and when I asked him about it, he said he didn’t need any. Go away, punk. Nurse Gave Alfie Evans Four Unidentified Drugs. Two Hours Later He Died. A new report in an Italian newspaper claims that a nurse gave Alfie Evans four unidentified drugs just two hours before his death. The newspaper indicates Alfie’s vital signs were strong in the hours before his death but that the nurse gave him four drugs while his father Tom was out of the room and his mother Kate was half asleep. Another family member was in the room at the time. So, to silence the press, the hospital promised Thomas more oxygen and more vital support. Two hours before dying the oxygen saturation was about 98 and Alfie’s beats were about 160, so much so that Thomas was convinced that they would let him go home soon (as the hospital administration told him on Friday afternoon) . Before dying, while Thomas had gone out for a moment, leaving Kate half-awake and another family member in the room, a nurse entered and explained that he would give the baby four drugs (no one knows what) to treat him. After about 30 minutes the saturation dropped to 15. Two hours later Alfie was dead. The report also interviews the head of the Italian hospital that had planned to take Alfie before British courts eviscerated the idea. The scientific director of the Bambin Gesù Hospital, Bruno Dalla Dallapiccola, is quoted as saying he thinks the lack of food and water during the first day after Alfie’s life support was removed contributed to his death as well as the lack of proper oxygen. Full text of Enoch Powell’s ‘Rivers of Blood’ speech, which was delivered to a Conservative Association meeting in Birmingham, England on April 20, 1968 — 50 years ago. Thank God For Making Me A White Guy! I REMEMBER THAT fine early fall day just like it was yesterday. My fishing pal and I were at one of the better trout rivers in my state, when we finally worked our way up to this one very beautiful, wide pool sitting beneath a verdant green cow pasture in the bright sunshine. We could just make out the pod of sleek brown trout silently finning in the middle current like a fleet of tiny submarines — here and there one would quickly surface to noisily suck down a struggling late summer insect drifting by on the sparkling, sunlit waters. A real-life vision of Elysium lay before us. Since it was my turn to take first crack, my buddy stayed up in the pasture to bird dog a little for me, or maybe to just laughingly narrate my angling mistakes to any nearby moo cow. Sure enough, my first two casts were pretty lame. But it didn’t seem to spook the fish much and on my third cast, I somehow managed a long, perfectly rolling arc, with the monofilament leader curving to the right like a major league pitcher’s best spitball — dropping my very own artfully tied grasshopper imitation just far enough upstream to realistically drift down into the trout’s kill zone. Whammo!  I lifted up my rod to gently set the hook and the four, maybe five pound fish danced and surged wildly across the pool. But I lost him after a thrilling half minute or so, when my gossamer thin two pound tippet gave way (which I should have changed out earlier). For whatever reasons, I always seem to remember every single big fishing battle I lost, instead of those all-too-few times when I do manage to land or boat a nice fish (usually released back into the water). But I guess that’s how us White people get better at things — we learn from our mistakes. Former East German STASI (secret police) Commie agent and White-hating Jewess, Anetta Kahane, was put in charge of censoring Whites on the Internet by Angela Merkel’s German government. The rats are working hard destroying White people. For years I rejected this conclusion. Even though everything I saw fully supported it. I remember when I first said to myself “it’s the only logical answer to all the BS going down,” my head swam.  I was just a regular joe, didn’t really “hate” anybody. A live and let live kind of guy — who liked drinking and fishing more than politics. But I could see certain things being quietly promoted in our country. The “Whiteys be bad” business and everyone else was “oppressed” by evil Whites. Funny, I couldn’t see any “oppression” going on and I definitely never got any special favors just for being White.  Notice how they’ve furiously been ramping up the “niggering” of America on TV in the last few years.  Hell, the brainwashing bastards are now casting blacks in historic White hero characters like Joan of Arc, Hannibal and Greek hero, Achilles. Jew Hollywood even felt free to turn Heimdall, Viking god of Nordic legend, black in a recent movie. Blacks are always cast in good guy roles — often the main hero these days. Recently, I saw a commercial for Ray Bradbury’s classic Sci-Fi story “Farenheit 451” where they turned the hero black. Movies also now feel free to cast black guys as the love interest for the White woman protagonists. White men must always now be the baddies in movies and the idiots in every TV commercial. Pay attention and you’ll see it all the time. A collection of tweets from Jews calling themselves “fellow White people” to trash the White race and lay down the guilt trip for diversity subversions. This is exactly the reason I say on my site (all the time) that even little Jews act in the destruction of the White race.  Ever since they came to America, the bastards have been busy brainwashing us White people — including in our schools. To demonstrate just how tirelessly our White Jewish brethren fight for Social Justice in this world full of hate, Ari Shekelstein has compiled these 26 tweets from the last 24 hours. I’ve seen this in real life many times. Big shot wannabees, who think they know everything and want to show just how great they are by doing things they know they can make someone else look at fault later on — should it turn out bad. In other words, they realize they’ll be the big heroes if everything goes as expected, but if it doesn’t they got an easy out. CYA. You can see how they mapped it out in their ambitious little pea brains. Last night we saw James Comey in a much hyped interview on ABC — one of the major mouthpieces of the NWO. And this is the 25th anniversary week for the Waco/Branch Davidian catastrophe. Very appropriate COHENcidence. In the case of Waco, those ATF fools could have easily nabbed David Koresh (another arrogant wacko) with a small team when he went into town, instead of doing a big, made for prime time bust on the main compound. ATF actually code-named the whole thing as “Operation Showtime,” if you can believe it. Hell, they already had an undercover embed and a team at a nearby house. A simple phone call would have been all it took.  You can tell it was a manufactured media event because ATF tipped-off reporters got lost and had to ask someone on the side of the road where the Branch Davidian compound was. As fate would have it, they just happened to be wacked Waco cultists who then alerted Brother man Koresh. The result was 4 ATF agents and 6 Branch Davidians dead, followed later by 76 burned alive — including many children.   Looking back over time, one can easily see them manipulating America. Israel has followed a clever, long-running geopolitical plan to reduce the surrounding Muslim nations into small, warring enclaves of Shia versus Sunnis. Funny how every single enemy of Israel and the Jews always ends up attacked by America eventually. Egypt and Jordan are bribed big time with our tax dollars (about 3 billion per year) not to do anything bad to Israel, so you can add that into your equations. In fact all sides – Bosnian Serbs, Bosnian Croats and Bosnian Muslims – were guilty of war crimes and indeed the Serbs were more sinned against than sinners. But the facts mustn’t be allowed to spoil the anti-Serb narrative. In a war few people understand, the Serbs were cast as the villains of the piece by Machiavellian Western governments and the duplicitous international media. And an ignorant, gullible public eager to have their thinking done for them, believed all they were told. Having fallen hook, line and sinker for the biased media coverage of the war, the demonized Serbs were a people found guilty as charged with no regard for the truth. I was reading over at Daily Stormer, the Internet’s most censored site, about some big Brit Naturalist White guy who was over in Africa trying to protect animals from poaching and he was stabbed to death by the natives. That’s when I remembered once kindly suggesting here on my site how it would be infinitely better to just let all Africans starve to death to help Mother Nature — instead of hindering her. Shocked?  But the White race has done more than enough for the black race by now. Instead of respect and love, they spit in our faces every minute of the day. Plus, they freely assault Whites on the streets with things like the “knockout game,” while robbing, raping, torturing and murder (including our elderly mothers and children). Most people don’t know the evil crimes blacks perpetrate against Whites, because the traitorous media works constantly to protect this criminal race. I’m not saying to keep them from eating food they get on their own from working for a honest living (lol) or from growing crops. Just stop giving them all that free taxpayer stuff and money on a silver platter anymore. If blacks want to stand up on their own two feet, that’s entirely up to them. We merely need to say no to the spoiled brat’s never-ending race demands and see what happens. So it seems like our capital has been riddled with foreign spies intercepting our leadership’s communications for years, and nobody has been bothering to do anything about it. Or at least, this is how this is being presented. But under the Trump administration’s leadership, it seems like the Department of Homeland Security is looking to crack down on all this “foreign” surveillance. The U.S. government has acknowledged the existence in Washington D.C. of what appear to be devices that could be used by foreign spies and criminals to track individual cellphones and intercept calls and messages, the Associated Press reported Tuesday. In a March 26 letter to Sen. Ron Wyden, D-Ore., the Department of Homeland Security admitted that it “has observed anomalous activity in the [Washington D.C. area] that appears to be consistent with International Mobile Subscriber Identity (IMSI) catchers.” DHS added that it had not determined the type of devices in use or who might have been operating them, nor did it say how many it detected or where. Just as Jewish academics lead the scholastic fight for egalitarianism in science and sociology, and Jewish media moguls lead the propaganda fight, the “civil-rights” movement itself found most of its leadership and financial support in the Jewish community. Almost from the first day of its inception in 1909, the National Association for the Advancement of Colored People (NAACP) was the premier organization working for a racially mixed American society. Interestingly enough, the founding board of directors had only one prominent Black, W. E. B. Dubois (who was actually a Mulatto). Most of the board consisted of Jewish Marxist ideologues. The U.S. House of Representatives and many state investigative bodies thoroughly documented the fact that all of the NAACP’s founders were activists in the Communist cause. Dubois even chose Communist Ghana as his burial site. The NAACP’s first president was Arthur Spingarn, and only Jews served as NAACP presidents from its founding until the 1970s. Noel Spingarn succeeded his brother, Arthur, and following him, Kivie Kaplan reigned over the organization. The Jewish leadership of the NAACP was little known by the public at large. When I came of age, the only name I heard associated with the NAACP was Roy Wilkins, who was its Black national secretary. Because he was so much in the press and public eye, like most Americans, I thought Wilkins was the NAACP leader. But Kaplan was the actual NAACP president during that time. Benjamin Hooks became the first Black president finally in the 1970s. Once a Black finally made it to the presidency of the organization, no longer did the public hear much about the NAACP “national secretary.” From then on the public spokesman was the NAACP president. THE POOR AND humble thronged in from all over the countryside, anxious to catch a glimpse of the Galilean called Yeshua. As the crowd pushed forward, they sometimes stumbled over each another — everyone excitedly hoped they might get close enough to witness another miracle, or just hear the man speak out His ringing words of Truth — words these people understood and felt so true deep down. The man called Yeshua saw all the people trying to get close. He knew he must soon speak out to settle them, or some may get hurt. In those distant times, the slightest injury could later mean a painful death by runaway infection. This hypocrisy is staring you in the face right this very day. Read on should you have any guts left and truly understand deep down what is right and what is wrong. Is Daily Stormer a Big Secret Jew Ops? Occasionally, I get comments in my moderation que from people saying Daily Stormer is a Jew operation. I leave them there, since I try to avoid starting up any business with “who’s a Jew” at my site. Sure, most of it is personality conflicts, jealousy — that kind of thing. I don’t have time for it. And the devious Jews use that weakness against us. Jews love to plant seeds of doubt in people’s heads among us real Awakened Whites. I got a book up on my shelf called “The Synagogue of Satan” by Andrew Carrington Hitchcock. In it, he claims Abraham Lincoln is a Jew. I don’t believe that’s true, but I do believe Lyndon Johnson was Jewish enough to be totally simpatico with the Jew agenda.  But the book is often too much — just how much I don’t know — so I don’t use it as a source with newbies. What I mean here is that a few awakened people think everyone is a secret Jew, but all the other regular “unawakened” folk (the masses) seem clueless on just how much Jews really do control things, especially the media. Yeah, I’m trying to look cool by making out my statement like it’s a big important quote by a big time writer guy. Jews are definitely not any big super heroes — as much as they like to imagine themselves. I remember once reading some Jews bragging with each other that the Jew race were like characters in a Harry Potter movie, while Goyim (non-Jews) were “Muggles” or whatever J. K. Rowling (a big lefty) called people who didn’t know about the secret sorcerer world and that hidden Hogwarts school, where they go to learn casting spells.  Barack Obama wants a million young clones of himself to promote Jewish Communism. Negroes tend to think very highly of themselves even though they have no redeeming intellectual qualities. We see this trait in Barack Obama. This Kenyan baboon is one of the most egotistical narcissists that we have ever seen in American politics. Here’s a perfect case in point. Have you had enough, White people? All the continual BS and lefty agenda promoted all the time? When the hypocrites now openly trash our race in the media — even the commercials? Stop and think: If you took out all black crime, the US would be one of the safest places on the planet. They are not going to say that on TV because of them protecting the criminal black race every minute. They’ve done it for decades. What we see going on today is a well-financed Globalist media effort promoting the kiddies to march for gun confiscation. Hell, they might even be secret, “Black Ops” false flag efforts behind many of these shootings. They obviously want to disarm White America, even though us Whites have always been pretty much a law-abiding race. Criminal blacks will always commit crimes on the vulnerable — guns or not. Whites will be the ones turning in guns, not blacks, believe me. Look at Great Britain. They disarmed Whites over there and then opened the floodgates to Third World mass immigration. Whites get raped, shot, stabbed, run over by Turd Worlders and Muzzies all the time. Whites get harassed, censored and even arrested for “wrongthink” now. Much the same in most European countries (Poland and Hungary have woken up). Us Whites need to organize for the coming Civil War. No doubt about it. Spell it out: WTFU! BARELY ANYTHING FOR THE BORDER WALL!   Pay attention and you, too, will see these lousy Jews everywhere telling America how to think. If only they could find a way to make underwater basket-weaving a thing to destroy the White race, they would be yapping away about that 24/7, too. Above is one Jill Wine-Banks on MSNBC’s “AM JOY” hosted on weekends by Joy-Ann Reid — a boot-lipped Sheboon who has zero problems insulting the White race, especially White men (us evil ones) — while doling out mega dollops of the lefty, Jewish-approved brainwashing.   Sometimes when I see one of these lefty creeps on TV, I take a quick cellphone shot. Then I look them up on the Internet to see if my suspicions of Jewyness pan out.  But I did stumble across daddy’s obit in the Chicago Tribune. Yep, he was a Jew accountant named Bert S. Wine, married to a “Sylvia” and they lived in totally Jewed-up Skokie, Illinois. I think it’s a safe bet she’s a member of the Tribe. And remember it was the Chicago Jew power nexus who brought us the wonderful Barack Hussein Obama. [VIDEO BELOW “continue reading”] That town has like 300 permanent residents, so they’re in for a pretty big dose of diversity. Let’s hope they don’t get racist about it. Isn’t it about time you started thinking about an Eirexit? Because otherwise, nothing’s gonna change. Blacks are basically semi-retarded jungle primates who love shiny objects, flashy clothes and drum beats, but who we unfortunately failed to ship back to Africa after the civil war like Mr. Lincoln wanted.    The only reason they are acting like that is because blacks are the subject.  I read over at DAILY STORMER about Michelle inviting this little chimplet (who we’re always expected to think as so cute or else you’re a racist) to her office after seeing a picture of the girl standing in front of that idiotic painting of Michelle at the National Portrait gallery. The two then danced around for the cameramen (called in beforehand to get PC propaganda material).  “Social media meltdown,” yeah, I bet. “Pure magic,” right. Ridiculous. You diversity-promoting liberal White people just don’t get it, do you? Our race is truly getting the wool pulled over our eyes. While the traitorous media constantly makes it out like us Whites are always the ones so evil and racist, criminal blacks horribly murder decent, innocent Whites on a daily basis.         AND THAT’S BEING NICE…    Good Thing Hitler Lost, Right? INCOG NOTE: I wanted to run Andrew Anglin’s piece here since I too feel the exact same way. I hope other Whites coming here will take a moment to read it and reflect on the matter.  Conservative cucks constantly signal about the “good war” against Adolf Hitler. But let’s take a step back here. If Hitler won the war, would this be happening? The Berlin Senate has paid for a 140-page teaching manual that provides instruction for teachers on how to teach gender diversity issues to pre-school children at nursery institutions. The educational initiative is called Queerformat. So far it is a first for Germany, but no doubt will soon spread elsewhere. Could The FBI Really be Controlled by Big Jewry? This question needs to be seriously asked since “Deep Staters” in DC are apparently getting away with conspiring against Trump before and after the election, while Hillary still lives richly and untroubled after all that has been uncovered about her in the last couple of years. AG Jeff Sessions has done virtually nothing, as the American public is continuously subjected to anti-Trump/Russian collusion BS night after night. Everyone is sick of it. Even the so-called FBI failures over Nikolas Cruz and the resulting shootings at that Parkland Florida high school should make one suspicious. My father himself once met J. Edger Hoover. Can’t exactly go into details, but he did. I remember asking him as a teenager if he believed the rumors about the Mob having photos of Hoover dressed in drag. Of course, the whole business of faggotry and tranny type stuff grossed both of us the hell out. But he couldn’t dismiss the possibility since a lot of people back then knew Hoover had a right hand man, Clyde Tolson, who never left his side and the two even vacationed together. Also, Hoover never married nor had any liaisons with broads that we know about to this day. Dad and I both used to read quite a bit about Mob history. I remember driving around Manhattan and surrounding NYC boroughs, spotting various locales famous in the world of crime, like Sparks Steakhouse where crime boss Paul Castellano and his driver Tommy Billati got rubbed out by “Teflon Don” Gotti’s boys in 1985.  And sure, my dad and I both knew that Jews were heavily involved in criminal Mob rackets, like Meyer Lansky — who died peacefully in his own bed after hiding out in Israel while setting the stage for an engineered acquittal in a Mob trial back in the states. But the idea of Jews controlling the Feds is truly the stuff of nightmares. I’m a Virginia boy, through and through. And I know the area where the woman was murdered. Hell, I know all corners of Virginia. This is generally a sleepy, quiet area, flat and very humid in the summer. Tobacco country. Along the highway 460 corridor, it’s full of lazy-ass, indolent and totally worthless blacks living on the government dole and selling a little weed on the side. But then again, same thing goes wherever the stinking apes live — city or country.  TerriLynn St. John was literally grabbed up Apeman-style from her own front yard in Wake, Virginia, in broad daylight, with her little babies in the house. Her broken necklace was found in the grass and her cell phone was off in the bushes. She fought back but the Apeman quickly overpowered her. My contacts tell me the traitorous media first suspected her White boyfriend did it after reading some of her social media posts where it sounded like she had a few problems in the relationship. This is why you had a brief mention on ABC “World News Tonight” of her going missing (local stations passed the word up to NY). This forced them to report Thursday night on the body being found, along with a quick shot of the black perp. But the story is now over. Gone. Finis. A black did it. No follow ups and involved stories later on “Dateline” or “20/20.” That’s only for White on White crimes. The woman might have had some relationship problems. The apeman could have heard about it, since he was acquainted with her (never, ever have contact with these beasts unless you have to). He decided to drop by and offer her his black greatness, but she told him to get lost.    You think it’s skinheads always getting us into never-ending wars in the Mideast? Oh, and I guess it’s the KKK constantly promoting Third World immigration into White countries and jacking up the homies all the time in the media? WAKE THE FLOCK UP, BROTHERS AND SISTERS! Amy Klobuchar: Jew Church Lady or Just Lefty? They had this lefty mouthpiece, Minnesota’s congress Democrat Amy Klobuchar, on ABC with Georgie Boy this past Sunday to talk about the shooting in Parkland Florida and also about “bots” dispatched across the Internet by the evil Rooskies “interfering in our democracy.” They love cramming in as much lefty and NWO geopolitical brainwashing into a segment as possible. Klobuchar said she wants media companies that don’t “purge bots” to be fined. In other words, they had better start vigorously censoring genuine people who don’t toe the company line of PC-dom and lefty “groupthink.” The “bots” crap is all a new scam to explain away awakening Whites commenting on the Internet and posting conservative opinions on Facebook and Youtube. I just know Klobuchar has got to be giant “CRYPTO” Jew.*  Plus, Klobuchar is always sucking up to the Jews and Israel, too. Hell, look how the state of Minnesota has been allowed to be infested with African Somalis, for chrissakes. Just stop and think: absolutely worthless criminal Somalis are imported into a cold weather White state, once a bastian of Swedish Christian White civilization — that alone should piss you off royally! So, I’ve decided to run a little poll (below) to see if you also think Klobuchar is Jewish. Whether she is or not, she may as well be one with the crap she’s pushing. War photog Bret K. Ellis took this now famous shot of captured Wakandan “King T’Challa” at the Battle of Atlanta. That’s first sergeant C. W. Moss escorting his majesty. As his men marched past the Leader standing up in the impromptu grandstand in company formation, four abreast, no two looked alike. They wore uniforms cobbled together from Realtree and Mossy Oak hunting garb to military surplus ACU and Woodland camo — along with a crazy assortment of ball caps, helmets and hunting boots. This oddball army might not look so fashionable but they could care less. Most carried generic AR-15s strung over the back with Magpul quick release slings or just any old found piece of strapping material.  Even a few Chinaman Norinco AK’s with those strange thumbhole stocks were to be seen. Many also had handguns at their hip or thigh, along with a nice dirk or combat hatchet. These were weapons brought from home, or scavenged off the dead on the field of battle. Most had some kind of military combat vest with magazine pouches, but others only had roughly camo spray-painted muslin Wholefoods grocery bags hung diagonally across the shoulder. Packed aboard all manner of contrivance came extra ammo, water, dehydrated prepper food and big canvas sacks of Russian spuds garnered by colluding with evil White guy, Vladimir Putin. Thanks, Bro Man! Fortunately, the 10,000 man First Cohort had a solid day of rest after moving up from Virginia along mountainous back roads. They marched by in good order — certainly no military academy graduates, but mostly in lock step. They were anxious to get where they were going and do what needed doing. As they stomped past the Leader in a muddy West Virginia beet field, they saluted with the right fist held downwards at a fierce angle and snapped determined faces up to look at him. They were digging it. “Coontact: I was grocery shopping today, the store was crowded and the lineups at the cash were long. There were several loudmouthed she-apes behind me in the line, loudly niggerbabbling at each other and their phones. Who Controls the Money? LOS ANGELES, CA.:  The faggot then dragged the poor kid back to a storage closet and sodomized him. Bald or shaved head, wearing a long-sleeve, yellow t-shirt, light blue jeans with designer tears and red Timberland boots with light tan colored soles. Anyone with info contact the LAPD’s Hollenbeck Division at |||PHONE\_NUMBER||| .
\end{enumerate}

\subsection{Berger's definition}

\begin{enumerate}
\item Aww, i'm flattered.   \&gt;on an internet forum.   That's where you are now, right? I don't understand your point. The whole demeaning angle you're going for kinda falls apart when you post things like "Blacks have no place in modern society. That's proven by the fact that they've never succeeded in a modern society anywhere, at any time. It's proven by the fact that every city they move into turns into a wasteland as Whites flee."   \&gt;Not an objective source.   He didn't ask for an objective source. He said "Would **you** say it's worth a buy now? Like at around 30 bucks or so?"  I'm not 100\% sure off the top of my head where I excuse "legitimate and valid criticisms of this game", but i'm sure you'll dig through my post history more and let me know.
\item Scapegoats Remember, it's the media's fault: Shameful. Several Senate Republicans denounced other lawmakers and the news media on Thursday for unfavorable depictions of the Iraq war and the Pentagon urged members of Congress to talk up military service to help ease a recruiting shortfall.When in doubt, see \#14 on this list. Any questions?. Shameful. 47 Comments: shameful is just one of the words for it. The ground-work is being laid to blame the loss on liberals and other critics at home. Just like they did with Vietnam. Well, perhaps I've got the wrong viewpoint, but sometimes I suspect it's the fact that so little of the current leadership has provided an example of servie that has a bad effect on recruiting. I wonder if Senator Inhofe is encouraging his children, grandchildren, nieces and nephews, or children of family friends to head down to the local recruiting station to sign up. Probably not. (That said, I'll cut Inhofe a little slack since he did serve as a Pfc in the Army from 1957-1958. But just a little.) These are the take responsibility people! If I only listen to Rush Limbaugh and watch Fox News and they are saying it stinks in Iraq, what now? When our government comes out with propaganda regarding Pat Tillman, Jessica Lynch, and they bring in their own news guy, Jeff Gannon, and those fake news report that stations were showing --- they went as far as having an actor pretend to be a reporter... well, it is a slap to the face to the American public. Those fake news report had no disclaimer and were meant to deceive us. What credibility do these guys have? They are the ones that want media consolidation. They are the ones that want PBS to be an extension of the White House. They are the ones that embedded the journalists in the war machine. They should be straight and just say what they mean, they want us to belly over and not think or speak. Great example of democracy guys. Karl Rove has a philosophy if something is said 5 times it must be true. So they'll just keep repeating it over and over; never mind the real problems. "because of all the negative media that's out there," Well, as a wise man once said, "you can't make chicken salad out of chicken shit." If they are so worried about missing recruitment goals they should be pissed off at those responsible for the piss-poor planning/implementation of the war and post-war occupation. It's a lot easier to send your child off to defend your country when you have full faith in the motivations, skills, and charater of those responsible for sending your child into harm's way. You want to fix the enlistment gap, work on those issues, don't worry about the media (everyone's favorite boogeyman)... Umm, call me crazy but I don't want my kids helping to form naked Iraqi's into pyramids; I don't want them driving in poorly armoured cars; I don't want them coming home with an arm or leg missing -- for a cause I don't believe in. If Bush is so hell bent on this war how come he didn't send Jenna and her sister? Well! is it not about time that we make some noise about recruiting Jenna and Barbara? let's see what they will have to say about that., or do they think these girls are priviledged? Their father was saying that enlisting is an honourable thing to do. why this would not be applied to his daughters, as well as all the young men and women of these senators who only know how to scold the public? Let's make a big fuss about this, may be they will shut up and start having a little understanding as to why it has been so difficult to recruit. Agree with the person who said the pubs are trying to set up a "liberals lost us Iraq" scenario. Libs didn't plan it, didn't have anything to do with managing it- but dagnabbit, if it wasn't for liberals the sunni insurgents and AQ crazies would have stopped planting bombs by now. Logical? No. But this is the right wing we're talking about. What is significant is that it is the Army (and, to a lesser extent, the Marines) that have recruiting shorfalls. The Air Force and Navy are having no trouble getting people to join. Might it have something to do with which service is most likely to get you killed? Forget Jenna \& Barbara. G.W. should be forced back into duty and be made to finish his national guard service driving a less than fully armored Hum-Vee in Iraq. Cheney, Wolfowitz, Rove, and the other neocons should be forced to drive the other vehicles in the convoy. They should be forced to stay on active duty until all other American forces have been withdrawn. Should they survive, they should be forced to live in exile for the remainder of their lives for the disgrace they have brought upon the memory of those who perished on 9/11. I was watching the last minutes of the Jim Lehrer Newshour last night, and they were showing the names/faces of the latest servicemembers to be killed in Iraq, and I couldn't help but notice, once again, how such a high percentage (almost all) of those KIA are from cities like Muscatine, IA; Monroe, LA; Laurel, MS; Pierre, SD. Places most Americans have never heard of, or been to. I live in a large city (Houston) and I can tell you, that the number of KIA from here is less than a handful (and almost all were non-WASP); and Houston is the nations 5th largest city. And I’m not sure, but I think the number of servicemembers in Iraq/Afghanistan, who are sons and daughters, of a Congressman is also less than a handful (it may be as little as 2). These percentages/ratio’s also hold up for the thousands upon thousand of servicemembers who have suffered very serious injuries, and have also been largely forgotten about. No need for me to repeat here, about how the Necon’s who pushed the War, have done everything in their power to make the KIA and injured nameless and forgotten, but the media’s complicity (with a few exceptions) in this silence is a disgrace. And, if you don’t believe me, just turn on FOX news (or CNN, or any national news show) at 12 noon today. The Air Force and Navy are having no trouble getting I don't know about the Navy, but during Vietnam if your number was close you joined the Air Force or National Guard, so you could have some semblance of control over life and a greater probability to live. Of course being in the National Guard these days appears to be the same as being a foot soldier. I saw the news the other night and they were again blaming the adults in the kids lives, parents, coaches, teachers. I thought that was hilarious. You mean parents actually care about their kids and watch over them; that seems like a very "family values" thing to do. I wonder if the recruiters are only tracking public school kids, or also kids in private schools; what about boarding schools. It is sad, the economic draft we have. My father came home from college one day in 1953 and saw a letter from the Army waiting for him. He turned around and went directly to the Navy recruiting center. He always was a smart guy . . . . As I read an opinion in USA Today (see snippet below), I got thinking. This man is a WWII veteran. He knows war is hell, absolutely necessary sometimes, and we need to support our troops (because war is hell). Yet, war is not the only patriotic action. Certainly demonstrating is, making voices heard, and having our press people be unbiased and report the facts on all sides. Yet this veteran (who happens to have found the paper) looks to others in the media to speak up, to report the truth, or at least give a voice to alternative views. Are we so afraid to say the President took us to war on a lie, that no good man or wowan can speak. Is fighting for our country worth it? And that means holding the President accountable. If the war was so right, why can't the President tell us the truth? I know there are a lot of people that value truth because they were very vocal about Clinton lying about sex in White House. My neighbor tells me this all the time. So I know truth matters. Bush lied, where is the outrage about truth now. I don't get it. I'm not a huge (just a little) fan of Senator Lindsey Graham, but I respect him immensely because he appears to have consistent principles and values, and he is a gentleman. The exact opposite is true for Cheney, Delay, Rove, Frist, Bush... snippet: ." Well, the analagy to 'Nam is playing out in this instance quite well unfortunately. Iraq is not Vietnam. There are many comparisons to find in every war. During the Vietnam era, we always compared our own revolution to the Vietnamese fight to get rid of us. Vietnam was never a threat to the U.S., while radical Islamists are a real threat. Ho Che Minh never promised to destroy America. while radical Islamists are a real threat I forget, what was Timothy McVeigh again... Cunning Realist did an excellent blog History's Rhyme about the similarities. I just finished the book “Losing Iraq: Inside the Postwar Reconstruction Fiasco.” and it was just really interesting to read about things that were known that were simply ignored (i.e., Bremmer terminating the military, so you had unemployed Iraqi's with weapons or the order was given that the only building to secure would be the oil ministry which added fuel to the impression that we were there for the oil. They say there were plans for the peace and exit strategy, but there is a certain faction in control that ignored them.). Report today that indicates the Army desertion rate for 2005 is already equal to all of 2004 (2,723). I suspect having your tour of duty changed and extended multiple times under that stop-loss policy doesn't help. Also don't you think, that soldiers that are coming home are talking to their friends and neighbors. I know the admin try to hide the injured and dead, but when there are so many, in so many communities, people talk. I think the Republicans should stop blaming others and fix the problems and be honest with the public. What if our goal is to own, at least have a very big foot in the door, the Middle East because of oil? Does that justify the war, and why can't the President tell us that? What was Cheney keeping secret in those energy task force meetings. If journalists aren't above the law, either should Vice Presidents. Not too shocking. Remember how the media towed the line for so long before finally beginning to ask questions just a month or two ago? This is just a scare tactic to try to get the media back on side, supporting the gov once more. This time, though, maybe they'll keep asking questions. Let's hope. Gee, I hate to break it to you all, but the facts are that the Tet Offensive was a failure...until Cronkite (an avowed Socialist) told the American people otherwise. This is a Holy War, you dumbasses. Why?! Because the people waging it on us SAY it is. The aggressor sets the rules, so what WE say about it does not matter one bit. If you think otherwise, too bad for you: You. Are. Wrong. And all the leftist press does is bleat on about quagmires and the noble, oppressed Muslim. Did you know that the Crusades were a RESPONSE to 400 years of fucking JIHAD?! Lying to kaafir is a commandment of Allah the Most Merciful to his faithful followers. If you believe what these Islamist douchebags tell us in English you're a bloody fool. "Religion of Peace" my NUTSACK. But you will see, in a few years, what I mean. Unfortunately, by then it will be far too late. Go read, idiots. This is a Holy War, you dumbasses. Hey Dumbass! Hope you're posting that comment from Iraq! And I hope you have all your dumbass brethern with you. Make sure you exterminate every last one of those unholy islamists -- 'cause that's what Jesus would want you to do! And when you come back home without a leg or an arm don't be expecting anything except a shitty little medal from the US gov't for doing your "patriotic" duty. Thanks in advance. Yikes. Who brainwashed that guy. while radical Islamists are a real threat You are quite right--however, the number of radical Islamists who were inside of Iraq has increased exponentially since we got there, not before...or did you miss that part? Not a Christian. Not an atheist. Not brainwashed. Not surprised. Get a clue. Or wait until France is felled from within by Islamists. Then there will be clues in abundance. boston I was looking for blogs referring to Dallas and found yours. Great blog! I'm in Dallas and here's some info about me Hi, I was surfing the internet and here I am at your blog. I'm quite impressed , with how you have put it all together. I'll be coming back again. Cheers, B.S. in Electrical Engineering LPN Online Degree Hello Friend i wanted your help in winning me the World's Biggest SEO Contest for v7ndotcom Elursrebmem and also the v7ndotcom Elursrebmem So can you please put my websites link in your website. This is a excellent blog. Keep it going. You may be interested in living better Southern California home loans visit Start planning for the future Everyone needs a good loan at some point in time. If you are looking for Long Island Roofing make sure have all the resources at your fingertips. The website 1stplaceroofing.com has a great informational directory on Long Island Roofing so you can find exactly what you need. credit center. Yours is top-notch! If you have a moment, please visit my site credit center I wish you all the \#\#name\#\#. Living in CA , we are always looking for ways to save us some money on any home in the south.line of credit if you know of a website or relaible business that can help us, please let us know.line of credit thanks Has the Housing Bubble popped? Boca Raton says "Yes!" A typical subdivision -- the Boca Country Club -- characterizes the area, and the problem. read more .... real estate for sale for sale Thanks for creating this blog, I like to cruise the blogs to see what other people think about. Regards, churchill dog home insurance nodding online Cyber-wandering and looking for something that might help my own real estate business. Stumbled across your blog and enjoyed the visit. Thanks for the read. Visit my site if you have a chance. Home Loans Student Loan Consolidation can help you reduce your interest burden by charging an interest rate lower than the rate on your existing loans. Debt consolidation loan can also allow you to make small monthly payments...... Signature viper alarm don't want yes men around me, i want people who will tell me the truth, even if it costs their job.viper alarm Hello Newbie here just looking for some info about - I just got an order of winstrall capsules from them. Cheers Great article! Thanks. Thanks for interesting article. Excellent website. Good work. Very useful. I will bookmark!
\item Internment: A Tool in the War on Terror? Michelle Malkin is a syndicated columnist and author of two books, of which her latest is In Defense of Internment (New York: Regnery Publishing, 2004). In it, she provides a defense of "threat profiling" already taken or contemplated since September 11. Ms. Malkin's earlier book was Invasion: How America Still Welcomes Terrorists, Criminals, and Other Foreign Menaces to Our Shores (New York: Regnery Publishing, 2002). Her syndicated column appears in nearly 200 papers nationwide. Ms. Malkin addressed the Middle East Forum in Philadelphia, on December 2, 2004. Millions of American schoolchildren have been taught that there was no evidence of ethnic Japanese disloyalty or espionage before, during, or after Pearl Harbor; that Franklin D. Roosevelt was hoodwinked by bigoted military leaders into taking draconian measures; and that the absence of espionage convictions of ethnic Japanese "proved" that the West Coast evacuation and relocation of ethnic Japanese was unnecessary and motivated primarily or exclusively by racism and wartime hysteria. My book debunks these deeply held myths regarding the moving of 112,000 ethnic Japanese from the West Coast of the United States to the interior of the country on Roosevelt's orders in February 1942. The order affected first-generation Japanese resident aliens and U.S.-born Japanese-Americans, as well as a relatively small but significant number of non-Japanese residents (non-citizens and citizens alike). "Internment" is actually a precise legal term for the centuries-old, worldwide practice of detaining non-naturalized immigrants during wartime. During World War I, some 6,300 European enemy aliens in America were interned by the War Department in prison barracks at Fort Oglethorpe, Georgia, Fort McPherson, Georgia, and Fort Douglas, Utah. During World War II, more than 31,000 enemy aliens from Axis nations were interned at Department of Justice camps—nearly half of whom were German or Italian. Civil liberties absolutists have invoked the so-called Japanese-American internment to attack virtually every homeland security initiative against Islamic terrorism taken by the Bush administration to protect America from murderous Islamic extremists. When the Justice Department asked Arab and Muslim foreigners to volunteer to help provide investigative leads, commentator Julianne Malveaux complained: "It's beginning to look like the Japanese internment." When two men were removed from a Continental Airlines flight in December 2001 based on the plane crew's security concerns, the ACLU compared it to "the Japanese internment issue." Even within the Bush administration, transportation secretary Norm Mineta (who was evacuated as a child to a relocation camp during World War II) has absolutely opposed any use of racial or ethnic profiling. Misguided guilt about the past continues to hamper our ability to prevent future terrorist attacks. We cannot win the War on Terror as long as we keep learning the wrong lessons about World War II. There are parallels between World War II and the War on Terror, but the anti-profilers and internment alarmists fail to make the proper comparisons. The Japanese espionage network and the Islamic terrorist network exploited many of the same immigration loopholes and relied on many of the same institutions to enter the country and insinuate themselves into the American mainstream. Members of both networks arrived here on student visas and religious visas. Both used spiritual centers—Buddhist churches for the Japanese, mosques for the Islamists—as central organizing points. Both used native-language newspapers to foment subversive tendencies. Both leaned on extensive ethnic- or religious-based fundraising groups for support—kais for the Japanese, Islamic charities for Middle Eastern terrorists. Both had operatives in the U.S. military. Both aggressively recruited American citizens as spies or saboteurs, especially (but not exclusively) inside their ethnic communities. Both were spearheaded by fanatics with an intense interest in biological and chemical weapons. The Roosevelt administration supported many national security measures—not just the West Coast evacuation and relocation—that took race, ethnicity, and nationality into account. Before Pearl Harbor, the secretary of war and the secretary of the navy informed the president "that they were instituting a program of employment discrimination on a racial basis whereby they would assure that future civil service vacancies in defense installations in Hawaii would be filled by ‘selected citizens of unquestionable loyalty rather than by citizens generally of alien extraction whose loyalty may be questionable.'" Ethnic Japanese were barred altogether from working within naval reservations. Immediately following the Pearl Harbor attack, about half of the nation's Nisei draftees were discharged amid espionage and sabotage concerns, no doubt because MAGIC messages had revealed Japanese spies had infiltrated the military. The rest were reassigned to noncombatant and non-sensitive duties. Soon after, the military stopped inducting Nisei. The creation of a segregated combat unit later permitted the demonstration of Nisei courage and loyalty to the world. That unit was barred from the Pacific theater of operations. With few exceptions, Nisei volunteers were prevented from choosing the branch of service they preferred. The navy, marines, coast guard, merchant marine, and air force for the most part did not accept Nisei into their ranks except on temporary duty as Military Intelligence Service specialists. With virtually no exceptions, Nisei soldiers were barred from participating in any cryptographic operations. It was prudent, while fighting a war against Japan, to apply heightened scrutiny to ethnic Japanese in the military. It is also prudent, while fighting a war against Muslim extremists, to apply special scrutiny to Muslims working in sensitive areas, including law enforcement, the prison system, and the armed forces. The Islamist infiltration of key institutions is as perilous now as the infiltration of Japanese loyalists would have been sixty years ago. America's military and civilian commanders made no apologies for putting security over diversity during World War II. Yet, out of fear of being labeled jackbooted racists, today's politically correct Pentagon and prison officials have rejected singling out Muslim soldiers and clerics for extra scrutiny. In the wake of September 11, savvy opponents of profiling have shifted away from arguing against it because it is "racist" to opposing it on alleged national security grounds. University of Toledo professor David Harris, for example, makes the bizarre case that allowing profiling based on race, ethnicity, or religion would "enlarge the suspect pool," requiring "all of those in the suspect category . . . to be stopped, questioned, searched, and investigated, even when their behavior would not have pointed to any reason to do this." But allowing airport security officials to take a passenger's ethnicity into account is not a mandate to stop every person of that ethnicity. Profiling is just one discretionary investigative tool among many to narrow the suspect pool, not to expand it. It is far from an infallible aid, but if law enforcement officials were permitted to use only foolproof techniques, they would be left with no tools to fight terrorism at all. As for the claim that profiling alienates minorities, the argument conveniently ignores post–September 11 polling that showed broad support among minorities, including Arab-Americans, for heightened scrutiny of those who appear to be of Middle Eastern descent. A poll conducted in October 2001 found that more than half of Arab-Americans agreed that law enforcement officials are justified in asking extra questions or conducting extra inspections of people who appear to be Middle Eastern. Separate polls showed strong support for racial profiling of Arab-Americans among blacks and other minority groups. On this issue, civil rights absolutists argue out of both sides of their mouths. They condemn homeland defense measures that use narrowly targeted criteria such as nationality (e.g., the special registration program, which required temporary visa holders from Muslim-dominated countries deemed to be of "elevated national security concern" to submit to fingerprinting, photographing, and stricter exit controls). Yet they complain just as bitterly when the government eschews narrow profiling policies in favor of broad security programs (such as the Computer-Assisted Passenger Prescreening System, for citizens and non-citizens alike, to be used at airports). They argue against racial profiling in favor of behavioral profiling. But when airport officials announced the adoption of behavioral profiling measures at Boston Logan (from which two of the September 11 terrorist flight crews departed), American Civil Liberties Union official Barry Steinhardt complained that it "likely will result in new forms of racial and ethnic profiling." Some might argue that profiling is so offensive to fundamental American values that it ought not be done even if it jeopardizes the nation's security. Yet many of the ethnic activists and civil liberties groups who object most strenuously to the use of racial, ethnic, religious, and nationality classifications during war strongly support the use of similar classifications in peacetime—to ensure "diversity" on college campuses, to guarantee business contracts for minorities, and to achieve socially engineered "parity" in police departments and public works projects. Encouraging public universities to assign "plus" factors to individuals according to their skin color is praiseworthy, in their view. But allowing an airport screener or consular official or deportation officer or FBI agent to assign "negative" factors on the same basis is a human rights abuse. The civil rights opponents of profiling have never met a "compelling government interest" for using racial, ethnicity, or nationality classifications they didn't like—except when that compelling interest happens to be the nation's very survival. We are at war with stateless enemies abroad and Islamist infiltrators at home who will not stop plotting to kill us unless we kill them first overseas and nab them preemptively on our own soil. Both my first book Invasion and my latest book arrive at the same conclusion: Political correctness is the handmaiden of terrorism. Related Topics: Counter-terrorism, History, Japan, War on terror | Michelle Malkin receive the latest by email: subscribe to the free mef mailing list This text may be reposted or forwarded so long as it is presented as an integral whole with complete and accurate information provided about its author, date, place of publication, and original URL.
\item The Trio of \#1 threats to our national security. This sworn enemy of all ‘infidels’ by which it means, anyone and everyone who has not succumbed and submitted to the totality of the shariah-compliant laws that govern every aspect of the Islamic way of life – is commanded by tenets of sharih law, to do violence including murder, to those whom they consider, ‘infidels’. This is the kind of off-charts insanity we are faced with when we confront this rapidly metastasizing cancer. Islam uses outright lies, deception, bribery, intimidation and violence to achieve its objectives. These tactics have proved effective in enabling their progress toward infiltrating and therefore, influencing all aspects of America daily lives. Islam is quickly becoming an increasingly influential force in the control and conduct of all our governmental, military and civilian institutions.  Are we all wearing blinders, or what? Number two in this triumvirate of aberrant anomalies that comprise the \#1 threat to our national security – is the current President, his entire administration and the Democrat majority leadership in the Senate and the Democrat minority leadership in the House of Representatives. Plus, the two most recent Jurists to sit on the Supreme Court. This conglomerate of corrupt ideologues centered in Washington, DC, believes that the major impediment to achieving their objective of affecting a total transformation of the basic structure of our Government, is the Constitution of the United States. They openly acknowledge their disdain of this unique document that was created by the genius of the Founding Fathers of this nation.  They choose to ignore, or circumvent or run roughshod over our Constitution, that most magnificent master plan that has served so well in governing this nation of free people, for more than two centuries since its founding. The result is a tsunami of regulations, edicts, executive orders and an endless litany of what I believe to be unconstitutional mandates that are being jammed down the throats of our citizens. All, without the slightest regard for our traditional institutions, established legislative practices, historical precedent or the tyrannical trampling on our freedoms – freedoms that are guaranteed under The Constitution. We are being led toward a disastrous demise by this current President and his mindless minions and enablers. Number three culprit in this trio of the \#1 threats to our national security, is what is popularly referred to as, the Mainstream Media. This amalgam of TV \& radio outlets plus, the mass print media (periodicals and newspapers) is slavishly beholden and embarrassingly subservient to Obama and this administration. And as such, they are complicit in enabling this president and his administration to achieve remarkable progress in the incremental destruction of our system of government. They do this by selectively reporting on significant events, by slanted coverage of the news, and by omitting or skewing vital facts that are integral to the news. This is by any measure, a dereliction of duty on behalf of the free press in the performance of their traditional role as monitors of elected government. That role calls for the ‘fourth estate’ (press/mass media) to delve into the actions of elected representatives, to seek out the truth and to report the entirety of their findings in a factual, unbiased manner. Thanks to traditional TV networks plus satellite and cable technology, we now have the capability of viewing news broadcasts from around the globe, 24/7. The reality is that this massive flow of news is funneled through a relatively few news bureaus agencies, a large percentage of which are little more than propaganda outlets for governments or huge international conglomerates. Thus, much of the information (i.e.,news) that winds up on our TV tubes and in the print media, is highly filtered and dominated by strictly controlled, agenda-driven special-interest overlords. A Johny-come-lately addition to the mélange of information is the burgeoning world of blogsites, a phenomenon that puts scant effort into vetting original sources or bothering to verify facts and figures. Thus, we have this flood of rumor, innuendo, inexpert opinion and quasi-information spewing onto our screens via the social networks and YouTube look alikes \& sound alikes. Very little of this volume of Ethernet space junk is ever validated but, that doesn’t preclude it from being multi-plexed and booted along by insatiable e-mail forwarders. Of course this too, contributes to the overall chaos and adds a further threat to our national security. And so, we come to the inevitable question used by so many people as an excuse for inaction: “What can I do about it?” The answer lies in an axiom taught by leaders in the military: “When you are caught in an ambush, DO SOMETHING!” Inaction in the face of the threats to our national security leads to certain defeat. Therefore, it is incumbent upon each of us to become informed. We are obliged to do the uncomfortable. It is time to do the unthinkable. This is the time to ‘educate ourselves’ to the \#1 threat to our national security.  Get involved and do something! The biggest threat we face today is the prospect that Obama gets 4 more years to screw up our economy even more than he has done so far. Never has a president had such an adverse effect upon the everyday lives of its citizens as this fraud we have in the White House. Instead of apologizing to the Afghans, Obama should be apologizing to us for the insane policies he has tried to shove down our throats. Noel Brand said... One of the biggest problems facing the Republican candidates is that they go on shows with Democrat partisan hosts who will question them trying to put them in a bad light. Picking and choosing the right venue is important for all Republican candidates otherwise they'll leave themselves open to getting involved in areas that they don't want to go. So Republicans, beware of wolves in sheep clothing especially on hostile broadcast stations.
\item With the persistent pursuit for gender equality, women have transcended the patriarchal norm that a woman’s place is in the sanctity of home. With the advance of capitalism, women have entered new arena where their capability, vitality and intellect are recognized or rather harnessed. Yet as women toilers in factories and business establishments, they continue to experience the same degree, if not greater, of discrimination and exploitation. Raised in a poor family who eked out a living from peddling food stuff for snacks in their barrio, Pola managed to finish high school but failed to pursue her dream of a college degree. Instead, she enrolled in a two-year course in a vocational school through the government’s “study-now-pay-later program”. In that so-called dual training, their only claim to being a student was the ID issued to them. They didn’t have a permanent classroom to pursue formal studies. Perhaps there really was no need as all they were taught to familiarize with different materials–wires, connectors and how to tape them together to assemble the harness of a vehicle. All they were taught were companies’ business concerns. In a semi-feudal society that served as mere supplier of semi-finished products to transnational corporations, perhaps those were all they need to know. After three months, Pola and her classmates were sent to a factory for on-the-job training as part of the course. They were supposed to be student trainees yet they were made to work like regular workers as relievers or substitutes to absentees. They received P240 per day’s work, part of which went to payment of their tuition fees. The remaining one and a half years of the course were spent in the factory with such meager pay and without any benefit, not even the mandatory social security for workers. Despite the rigor of the job, Pola worked hard, patiently waiting for the training to end in the hope that she would be taken in as apprentice. She got the job, true, but it did not take long before she was laid off. Thus began Pola’s rollercoaster journey into the world of commodity labor, exacerbated by the onslaught of imperialism’s neoliberal globalization as it dashed fumbling for a panacea to its crisis. The woman’s values of good-naturedness, patience and subservience inculcated by a feudal class society were fully taken advantage of. Pola later applied as a saleslady in a well-known mall in their province. But she resigned after a month. She could not stand the difficult working condition and the ridiculous and repressive policy of the establishment. For a measly wage, she had to remain standing the whole day to reach her quota for the brand of dress apparels she was selling. There was a time when she was reprimanded for bringing her handkerchief inside the store without first registering it. Personal belongings had to be registered before bringing them in lest you would be accused of stealing. From the job in the mall, Pola worked in an electronics company where she assembled “male” and “female” terminals used in television sets. But after more or less four months, her contract ended. This was the endo (end of contract) they call in the labor lingo. Pola ended up in a food factory, where she was hired through an agency. With a spoon, she raced after the cups of noodles to determine if the noodles and condiments were of the right quantity or if needed to be reduced, add on, or changed. Also, if the machine that put on the cup lids was out of order, she had to do it manually. They worked by shifts in the factory. There were three shifts in all. But if a worker for the next shift was absent, she was obliged to take over and work up to 16 hours. Then again, it was endo after five months. Pola also tried working as caddie in a golf course. She was an umbrella girl who trod on the heels of the golfer to shed him from the sunlight. But unable to stand the harassment from her bosses, she left the job after two months. Through an employment agency in Makati, Pola was back as a factory worker. This time it was in a company manufacturing plastic lids for bottles of lotions, medicines, etc. Initially, her job was trimming the extra plastic around the lids to even them out; later, she was transferred to the packaging section. Sometimes, she relieved the operator of the machine that molds the lids. As trimmer her quota was 6,000 plastic lids a day. Due to the thinness of the lids and the absence of a protective devise, her fingers often got wounded. As instant remedy, she would put on some adhesive tapes. But in the long run, her fingers have become numbed that she would not mind at all anymore. If she had not reached her quota, she was obliged to go on “overtime-thank you”, meaning overtime without pay. Again, after five months, endo. But she could continue working there as an “extra”– doing the same work, but with lower pay and without a contract. Since life is difficult for Pola, any job is a welcomed treat just to earn a living. One day, coming home from an arduous day’s work in the factory, Pola met some students who stayed in their community. She was invited to sit-in to their discussions on the Philippine society and revolution. That awakened her to the stark realities–the immense oppression and exploitation of workers like her, as well as of peasants, professionals, youth, women and other sectors in society. She learned that their affliction was not destined. It was designed–a sinister scheme of the ruling class to hold on to power and wealth. But the greatest lesson she learned from their discussions was the solution to the people’s problems. Pola could not contain her rage, as well as anxiety, with that realization. All along she had been entertaining the thought of leaving her job in the factory which did nothing but extract the workers life blood and sinew to accumulate huge profits for the capitalists. After thinking it over for days, weeks, and on to several months, Pola finally decided to work full time in the movement. This was the most decisive action she took in her whole life. She has the chance now to look at life from a different perspective and open up to new opportunities, best opportunities. Sometimes, she reminisced about her past life in the factory, in the mall, in the golf course and how she spent it in vain. She could do nothing about it now but it would serve as a potent inspiration for her to get involved and take action to change this oppressive, unjust structure. After more than a year of working in an urban center, Pola is now Ka Lina, a red warrior of the New People’s Army. She no longer held spoons, wires, connectors, dresses, umbrellas or plastic lids. She now carries an armalite. The broad countryside is her school and each day they delve into the strategies of the people’s war that will topple the semi-colonial, semi-feudal structures that oppress the people.
\item Ali Belaroussi and his colleague were kidnapped last Thursday "The head of the Algerian mission Ali Belaroussi and the diplomat Azzedine Belkadi, whose government is ruling in violation of God's will, were killed," said the written statement on Wednesday, which could not be independently verified. Belaroussi and Belkadi were kidnapped at gunpoint last Thursday in Baghdad's upscale Mansour area. The pair appeared in a video made public on Tuesday blindfolded and in captivity, giving their names and home addresses. It was the first time they had been seen since being abducted. Wednesday's statement, which appeared on a website, claimed the envoys had been killed because of the Algerian government's repression of Muslims in their North African country. Al-Qaida blames Algeria"We won't forget what Algeria did to Muslims, by killings, destruction and spilling their blood," said the statement. Al-Qaida in Iraq, the group led by Jordanian Abu Musab al-Zarqawi, has claimed responsibility for attacking three other diplomats from Muslim nations. Azzedine Belkadi appeared in a video made public on Tuesday Egyptian envoy Ihab al-Sherif, 51, was seized on 2 July in Iraq as part of an apparent campaign to undermine Arab nations' support for the Iraqi government. Al-Qaida in Iraq later claimed al-Sherif had been killed, but provided no evidence and his body has not been found. After al-Sherif's kidnapping, attackers in Iraq fired on envoys from Pakistan and Bahrain in what police said were kidnap attempts. The Pakistani escaped unharmed, and the Bahraini envoy was slightly wounded. SOURCE: AFP
\item Its lyrics call upon "soldiers of Allah" to "crush our enemies" and celebrates "martyrdom" as a glorious victory. The chant - known as a nasheed - was played during Link FM's breakfast show on December 16 and 22 last year. Ofcom launched an investigation after listeners complained that it "promoted terrorism" and contained "Jihadi lyrics". It found: "We considered this Nasheed to be promoting a narrative closely associated with propaganda used by extremist organisations to recruit members. "This Nasheed would have been understood by Arabic speaking listeners as being an implicit call to action to encourage people to join a form of violent Jihad. Ofcom said the radio station was in breach of broadcasting standards and is now considering statutory sanctions against it. The sanctions include a correction to be aired or a financial penalty. They could even revoke its licence and shut it down for good. The breakfast show's presenter said she did not speak Arabic and did not understand the chant before playing it. Nor was it vetted, the Ofcom investigation found. She said she was "extremely sorry and horrified". In a letter she added: "I sincerely apologise and feel quite embarrassed about the reputational damage to the station and myself this has caused." The Pakistan Muslim Centre - which owns the radio station's licence - said it "wholeheartedly apologised for the error"
\item It's time to imagine a world without Islam. I was at a meeting, where the final speaker said, “Muslims are okay guys, they go to church just like we do.” Muslims don't go to church, they go to a Mosque. People tend to believe that Muslims are wonderful people and have too much “Christian Character” to engage in atrocities. In 2001, after showing a Christian lady the Voice of the Martyrs magazine, she kept saying, “They can't do that, that isn't Christian.” I had a hard time convincing her that Muslims aren't Christian, in spite of all the evidence I could present to her. Her incredulity notwithstanding, Muslims aren't Christian; they exterminate Christians. Islam was out to conquer the world since 622AD, long before America was founded. No nation has ever willfully accepted Islam, it was imposed by violence upon them, including in Saudi Arabia and Iran. Bush 43 called Islam, “a religion of peace”. This lie is in the translation. The name Islam can be translated two ways. It can mean “peace”. “Islam” is an active form of the Arabic word “Salaam”, meaning “Peace”. For example, “ Dar Al Islam” means “ The House of Peace”; and the “Dar Al Harb” means “The House of War.” Islam also can mean submission, or slavery. Slavery continues to this day in Islamic nations, even though Saudi Arabia and Yemen banning the practice, on paper in 1962, and Mauritania doing the same thing in 1980. A singular definition of the word Islam would be, “peace equals slavery.” The rest of the world's definition of the word peace means the, “absence of violence and “serenity ”: the Islamic definition of “peace” means slavery. “Fight those who believe not in Allah nor in the last day.... Nor acknowledge the religion of truth, even if they are people of the Book, until they pay Jizya with willing submission, and feel themselves subdued.” The Koran, Surah 9:29. There are over 100 Jihad verses in the Koran. You cannot make peace with men who have such a mindset. While the average Muslim doesn't fly airplanes, they content themselves to pouring out of their Mosques and hacking normal people to death with machetes. They form mobs, break into Churches and stab Pastors to death. This has been Islamic behavior for 1400 years. These behaviors are commanded in the Koran and Haddith. So is Jihad. Jihad isn't just violent words, it is backed up with deeds, real and actual. The “prophet” Mohammad engaged in over 40 battles, raids, and ambushes. The Hadith recorded the names of 27 people he had assassinated. He had hundreds of innocent people beheaded, including over 600 Jewish men, not for crimes, but for not worshiping Allah, and accepting Mohamed as a “prophet”. Thomas Jefferson authorized the Marine Corps to sack Tripoli, because the Muslims were enslaving our citizens.  Castration was the usual lot of all slaves. After studying the Koran, Thomas Jefferson knew he couldn't negotiate with the Ottoman Turks, because they were Muslim. That is where “The Shores of Tripoli” came from in the Marine Corps Hymn. In WWII, Adolf Hitler had regiments in the SS that were Islamic. February 14, 1979: Muslims occupied the US Embassy in Tehran, Iran wounding one US Marine, who was held hostage by the Khomeini government until February 21. November 4, 1979: our Embassy staff in Tehran was held hostage for 444 days; by Muslim “students”. April 27, 1980: the bodies of 8 of our servicemen were openly cursed, by Muslim Imams; after they died trying to extract our people. April 18, 1983: Muslims attacked the US Embassy in Beirut killing 63 people. October 23, 1983: Muslims attack US Marine Battalion HQ, killing 241 people. February 26,1993: Muslims first attack on the WTC. 6 people killed. In October 1993, we lost 18 lives at the hands of Somalian Muslims in Mogadishu. August 7, 1998 US Embassies were bombed in Kenya and Tanzania, killing over 260 people. In October 2000, the Muslims attacked the USS Cole. 17 sailors died. 9/11/2001, the worst Act of War ever perpetrated on the American people. 2,976 people dead, and over 6,000 injured. The whole Uma ( Body of Islam ) congratulated themselves, thus incriminating themselves. This is a small portion of Islam's history. Conservatives will probably say, “How does this affect American interests, and we aren't the guarantors of foreign independence.” That is a good question which I shall answer. Any threat to take over the World is a threat to American Sovereignty and American security interests. America is still on the Planet Earth, and not on the moon. Never forget, that if we lose our Nation's Sovereignty, we lose our Christian Law System. The Muslim wants to replace Bible based Law with Koran based Sharia Law. America, being Christian; is part of the “Dar Al Harb”, which is slated to be consumed by the “Dar Al Islam”. Islam isn't the only enemy seeking America's destruction that we currently have. They are the only enemy that is on the march right now. Islam is the only enemy that has attacked us physically. The North Korean wishes to destroy the US, but they have not attacked us, so for now they should be dealt with in a low-key manner. One cannot say the same thing about Muslims. Muslims only understand force. Even our surgical strikes are lost on their ape-like minds when they retort, “Allah has cursed their weapons that is why so few have died.” The fact is, the Muslims commit acts of Terrorism because they are Muslims. Terrorism is only one weapon in the Islamic arsenal. Terrorism is modern piracy. Therefore, our “War on Terror” is a farce. In 2001, I outlined a Victory Plan that I formulated back in 1984. It has nothing to do with behavior modification or standing occupation. It has everything to with destroying our enemies. Christians should step up their efforts in prayer, preaching, and revival for America. If we are not right with God, he will not bless our nation. Our Churches should realize that they cannot covenant with Muslims. Not only is yoking with Islam ecumenical idolatry, it is suicidal. It would make sense to change our Nation's economic and energy policies so we can drill American oil for the American people. Why should we enrich people who are out to kill us? Terrorists should be treated as pirates, and summarily hanged or shot. Congress should issue letters of Marque and Reprisal to any group that wishes to exterminate these pirates. Congress should lawfully declare War on Saudi Arabia. Saudi Arabia is the “fatherland” of Islam.  Our incursions into Afghanistan and Iraq were playing around the edges. Mecca should be destroyed, and the Ka'aba stone blown apart, and film it. Mecca is the Capital of Islam. We are in a War of competing ideas. The Ka'aba is an embodiment of Allah. The Muslim needs to be shown there is no Allah. Allah is not God. We ought to make war on Islam; they have been making war on America before America was founded. They have been at war with the world since 632 AD. New governments should be instituted with Constitutions that embody the Christian Law System, as opposed to the fantasy of “Islamic Democracy.” (Iran has a Parliament, and Iraq’s democracy isn’t working.) This would be done by the people themselves, not by America. After we destroy our enemy, we should not engage in Nation Building. The Church can send missionaries and build nations that way. Now the woolly minded will be concerned about the pride, well-being, and rights of Muslims. We live in an age where people's only concern is for the welfare of the guilty. Americans need to get in tune with the fact that civil authority exists to punish malefactors. I'm not talking about a “Catholic Crusade,” I am talking about an actual War. The last war America fought as a war was WWII. We should treat Islam the same way we treated the NAZI Party. If we don't start thinking clearly, we will have a Muslim ruled America, sooner than later. If we don't put Victory on the table, and Destroy Islam, we will be slaves in an Islamic world. If you love God, and then your Country, you should be willing to contend with the enemies of God.  After 1400 years, it is time for Islam to go.
\item January 23, 2016 (KHARTOUM) - Sudanese government and Sudan People’s Liberation Movement - North (SPLM-N) delegations Saturday concluded their second informal meeting with no sign of progress toward an agreement ending the four year’s conflict in the Two Areas. Instead, the two warring sides exchanged bitter accusations. Khartoum said the rebels have drawn back from previous understandings while the SPLM-N accused the government of refusing to make the needed concession for the sake of peace. When the two delegations began the second informal discussions in Berlin on Friday, there were great expectations that they would end the meeting with a positive note, opening the door for the signing of two deals at least on the cessation of hostilities and the humanitarian assistance in order to pave the way for a pre-dialogue meeting. However on Saturday, the head of the government delegation Presidential Assistant Ibrahim Mahmoud Hamid issued a statement saying the SPLM-N delegation pulled back from previous commitments to join the national dialogue conference in Khartoum. Hamid further said the SPLM-N relinquished the principle of the unity of the national army, which was the base of an agreement reached during the first informal meeting in Addis Ababa last December. "The Government of Sudan regrets this position, which represents a setback from the previous understandings that established principles to achieve a comprehensive peace. Also, it frustrates the aspirations of the citizens of the two area and the whole people of Sudan. " Following a meeting of its leadership from 14 to 17 January, the SPLM-N released a long statement on Wednesday 20 January, dealing among other with informal discussions in Germany. The SPLM-N stressed its commitment to a comprehensive solution and the self-rule for Blue Nile and South Kordofan states. Further, it pointed to the need to reach a new agreement on security arrangements to ensure the formation of one army after the implementation of the peace agreement. The statement welcomed the call of President Omer al-Bashir for a national dialogue but reiterated its rejection to join the ongoing conference held in Khartoum. The SPLM-N and the holdout opposition forces call to implement a number of confidence building measures aimed to create a conducive environment before they take part in the internal political process. In a short statement released on Saturday, the SPLM-N spokesperson Mubarak Ardol fired back at the senior Sudanese official, saying he refused all the propositions of his group to end the humanitarian crisis, hold an equal national dialogue, a simultaneous cessation of hostilities in Blue Nile and South Kordofan states and Darfur region, and to reach new political and security arrangements for the Two Areas. Ardol said he would issue a detailed statement about what happened during the second round of informal talks and stressed the SPLM-N commitment to "key issues for the Sudanese people related to peace, food, freedoms, and citizenship, without discrimination." Arman and Hamid briefed the African Union representative and the head of East Africa department at the German foreign ministry about the outcome of the meeting. Ardol said the second round is the last in the informal talks. On a related development, the head of Humanitarian Aid Commission (HAC) Commissioner Ahmed Mohamed Adam accused the SPLM-N of confusing the political and humanitarian actions. In a press conference held in Khartoum Saturday, Adam further urged the SPLM-N to abandon its demand to transport humanitarian aid to the needy in the rebel controlled areas in South Kordofan and Blue Nile through South Sudan and Ethiopia. He stressed the government will reject any relief comes from outside Sudan, and reaffirmed Sudan’s commitment deliver aid to the civilian in all the areas controlled by the SPLM-N. Since several years, the two warring parties accepted that UN agencies deliver humanitarian assistance to civilians in the war-affected areas. But Khartoum insists that the operation should be controlled by HAC and the rebel refuse saying this agency is infiltrated by Sudanese security agents. This was what i said from the beginning,there won,t be progress as we knew it was tactical to buy time to resume war,for the NCP to accept peace,the war should be taking side big towns like Elobied,Umrawaba,Khartoum,Kosti etc where real jalaby should also suffer and there are the one to put pressure on government to sign the Peace,See when SRF entire Abukorshola and Umrwaba,all Arabs were crying. How comes you denied humanitarians access for your people and you bombed them indiscriminate pretend are all Sudanese,i say no,enough is enough for all what you did to our grand!grand! fathers,no one should be fooled for fake peace agreement,our leaders should think how we can take the fighting inside Khartoum,international community will never assist us,we make it for our own,Alatua Continua!!
\item Prime Minister Binyamin Netanyahu, even after failing to form a government, may opt for a new election rather than let his enemies win. That way, the decision reverts to the voting public rather than the politicians. Netanyahu has had an amazing run, leading four governments at the head of a successful Likud. This time, he is making a dangerous personal gamble. He may calculate that even if some voters make him pay for failing, the outcome is unlikely to be radically different from the April 9 results. The right-wing bloc which he leads may even gain by sweeping up the 300,000 votes that were lost in April when New Right and rightist splinter groups failed to make it past the threshold. The question is who will benefit? Netanyahu is beset with legions of rivals and foes, some on home ground in Likud. He faces another harsh test of endurance in which they will go all-out to prove that his failure to form a government shows he has lost his celebrated political touch and it’s time for him to go. Avigdor Lieberman, who parlayed his five-member Yisrael Beitenu into a club for denying Netanyahu’s 35-strong Likud a majority government, will most likely survive. The only substantial change to be expected is the decline of Blue-White, the opposition grouping born shortly before the April election and which came close to beating Likud. The next time round, Blue-White will have lost its luster as a new alternative for displacing long-running politicians. It is moreover top-heavy with four leaders – Benny Gantz, Yair Lapid, Gaby Ashkenazi and Moshe Ya’alon – whose entire repertoire consists of one song. Binyamin Netanyahu must go! Even Madonna’s audience demands more variety. This quartet have tried signaling the Likud party: Get rid of your leader and we can get together for a unity government. Rightist or leftist politics or any other ideology don’t seem to matter, just the removal of the major impediment to their goal. That is just one of this opposition party’s mistakes. Their cherished slogan may work for the individuals, groups and media dedicated to toppling Netanyahu, but have the reverse effect on the street where the voting public is to be found. There, the long-running hate campaign against Netanyahu only boosts his credibility as a nationalist, populist, charismatic leader. This matches the current trend sweeping many other leaders to power in the West – not only in the US, but in India, Brazil, Germany, France, Hungary, Italy and other countries. On Saturday night, Blue-White staged an anti-Netanyahu protest rally in Tel Aviv. At the last minute, Gantz invited MK Ayman Odeh, head of the Israeli Arab party list, to join the speakers. He committed the political blunder of giving an Arab nationalist ideologue equal footing with Israeli nationalism, in the hope of drawing a mass audience. This blunder will haunt him in the forthcoming election campaign for winning the votes of the Jewish majority. So will another: Blue-White orators have joined the prosecution’s campaign, backed by former judges and legal buffs, to force the attorney general to indict him on charges of corruption. Yair Lapid vociferously accuses him of attempting to engineer parliamentary immunity to evade a trial. These opponents present themselves as the guardians of Israeli democracy. The prime minister claims he and his close family are victims of a political witch hunt. The general public is not entirely convinced of his guilt, especially when he is loudly condemned by “left-wing” politicians and media. Enough voters may be willing to give Netanyahu the benefit of the doubt, so long as he is not proven guilty in a court of law. By the same token, many will turn their backs on his accusers. The prime minister went on the air after a 22-minute meeting with Lieberman failed toto break through the latter’s refusal to make concessions on the draft law obligating the drafting of yeshiva students to the military according to agreed quotas. The ultra-Orthodox parties, which have fought against the measure, gave way on certain points to smooth the negotiations for a right-wing national government. Lieberman’s five-member faction holds the key to a Knesset majority of 65 versus the opposition’s 55 members. In his speech, Netanyahu said: “A month-and-a half ago, the people spoke and ordered us to set up a right-wing government led by myself. I have made tremendous efforts to obey the people’s will by setting up this government and averting another superfluous election.” He went on to say: “There is no reason in the world to put the country on hold for another year and-a-half and squander billions. Unfortunately, up until now I have not been able to persuade Lieberman to prevent this vote taking place. Earlier on Monday, a 65-majority of the Knesset, less than six weeks in office, gave the bill for dissolving itself preliminary and first readings. There were abstentions. If Netanyahu fails to form a government by the Wednesday deadline, he intends to let the measure go through to second and third readings, clearing the way for new elections at the end of summer. The alternative dates cited are Aug 27 or September 3. You promised Israeli law for Judea and Samaria if your right wing bloc was elected. If you would have done that first, your coalition would very likely have united around you. You didn’t really intend to keep that promise did you? Why are you so much against Trump, eh? Repent right now, and forgiveness is available for free, but it has to happen before June. After June, it might be that this is not a sin no-more. where in the bible does it say this? Get your acts together and re-make your political parties into something more effective and less partisan. I think that Israel should stop ripping itself apart with nit-picking partisan small-brained politics and come together to avoid an election. There are more important stuff than minor differences blown out of proportion to gain power.  You know Israel will never drop a nuclear bomb first. But if you try, you will get a nuclear bomb right on your nose. No shouts required and no doubts about it, the Lion Judah from Philadephia! Designed to best fit the given situation. This is the episode S08E07 director’s cut from Game of Thrones. The royal jester Avigdor locks himself in the mensroom and blocks the door from inside, in the only john of the castle. People with john-needs serenade at the door dreaming to get in. He might open, but we do not yet know for sure. Just give in. Fat yeshiva boys dont want to fight. They sit and learn in yeshiva. You dont need them in idf. Helping out in religious shools or housholds should be an obligation for them in order to give back to society. So they can serve the people if they dont want to serve in IDF. Ridiculous stall by a former Russian Jew if my memory serves me. Not bad for a transgender, eh? So the public elects the PM and the Israeli government who doesn’t like the majority’s decision gets to grind things to a stop. What kind of F-cking government is that? Do all Turkish take it up the tuches like erdogan? Or just the pisliklers that post on DEBKA…? Even if Bibi doesn’t prevail this time around, he may be back for yet another round in the near future. Already done that, I believe. I’m sure Israel will survive long enough to see him re-elected, if that is in his future. The Trump Impeachment is BECAUSE of the Cratering Netanyahu - SNAP! The Trump Impeachment is BECAUSE of the Cratering Netanyahu – SNAP! i see we have a turkish vermin infestation…. -Anne karnında amca yarrağına doymuş orospu çocuğu seni.
\end{enumerate}

\subsection{Schmid's definition}

\begin{enumerate}
\item Paul is the eternal jew par excellence, to borrow from Nietzsche. Replace “Paul” with “Auster” and voila you have a current Porphyry polemic. Works for Paul Gottfreid, too. Christian metaphysics are so primitive compared to their classical contemporaries. Anything healthful in Christianity was due to the White man’s efforts. I think this is why some WNs defend Jewsus. Teachings of a crucified rabbi are deliberately jumbled with White philosophy tacked on centuries later. I agree that the Paul comparison is applicable to the likes of Auster and, say, Takuan Seiyo. But Gottfried strikes me as the exceptional “good Jew” (I believe this is a term by William Pierce). Of course: Gottfried’s views will always be influenced by his Jewish background. But unlike Auster, Taksei and especially Fjordman, Gottfried is not hiding anything, on the contrary: he has been defamed by his fellow Jews. Listen for example the recent interview of Gottfried by Robert Stark at VOR. But as I said in the other thread on Nietzsche, I’ll try to post entries of high criticism about Christianity, as high as I can. And the truth is that theologians (think for example of Karl Barth) have made fools of themselves by taking Paul seriously through enormous, dense treatises of classical Christology. I believe that Hoffmann’s translation of Porphyry is essential reading for the scholar. I have done a modest research in immensely rich theological libraries in the US and can conclude that, even in fragments such as the literary remains of Porphyry, the sharp criticism of this philosopher who flourished in the 3rd century has more value than the mountains of exegesis I’ve found in theological libraries: an out-and-out fool’s errand that has lasted almost two millennia. Gottfreid may not be hiding anything. He could be honest. Certainly his Judaic background colors his views as they should. But, why is he worth our time? Do we need to hear his specific anecdotes to win? Our Judaic Question can be boiled down to several simple points. More pressing matters are at hand( the defiling of our women and land). Paul Gottfreid strikes me as a court Jew. A luxury at best and a hindrance at worst. Have you listened to Tanstaafl’s most recent VOR show? He and Caroline spend a few moments on him. I’m spending more time on Paul Gottfreid than he’s worth to make this point. A stitch in time saves nine, no? Whilst hunting for more Porphyry, I stumbled upon the works of Julian the “Apostate.” Hope you and other readers can benefit from this translated work. why is he worth our time? I am not saying that white nationalists should have Gottfried as a mentor. Only that non-nationalists who might be biased against us may find in Gottfried a first, preliminary step to later listen our more radical views. Have you listened to Tanstaafl’s most recent VOR show? I have a damn good Latin-Spanish translation of Julian’s book in my personal library and I am indeed planning to post a couple of entries about Julian (and Celsus too). Porphyry on Christianity, Nietzsche on Christianity, uncle Tom cobbly and all on Christianity.
\item Please help keep this website going by making a small donation in support of Leo’s work through Fundly: or Patreon: Article by Leo Zagami Italy, and Rome, in particular, was once known for the works of Caravaggio, Michelangelo, and Raffaello, beautiful works of art that gave Rome a reputation like no other, but that is no longer the case after the latest stunt by the Museo di Arte Contemporanea di Roma (Museum of Contemporary Art of Rome) that has specialized in Italian contemporary art from the 1960s-onwards. The museum is operated by left-wing idiots and has two locations, one in Via Nizza and the other in Testaccio. This weekend, at Via Nizza a blasphemous poster was displayed of Jesus with an erection in front of a child with the words “Ecce homo” and “Erectus” that the artist in question dedicated to the portrait of a pedophile Jesus. This scandalous representation was denounced by the politicians of Fratelli d’Italia, Fabrizio Ghera and Andrea De Priamo, and soon afterward it was removed. However, Rome is becoming increasingly heretic these days, from the procession of Pachamama to Moloch at the Colosseum to the Communist songs featured in a Sunday mass, and with the still ongoing activities of the pedophile priests, we can see how the Vatican is permitting and even promoting a mockery of the Christian Faith that will not be forgiven by God. For some people, it is hard to imagine that this generation might witness the destruction of Rome under Francis’ heretical pontificate, but his seven-year period of tribulation may end in March of 2020 with a big surprise for the Vatican degenerates and their city of sin. It was written by St. John the Evangelist over 2,000 years ago in the Apocalypse, the Third Secret and many other prophecies this would happen, but it was not until St. Malachi’s Prophecy of the Popes, that a time for the destruction was given to be at the end of Francis’ period of tribulation, that is coming up next year. Rome’s blasphemy deserves the attention of God and he will make them pay for their sins once and for “The Blasphemies are piling up in Rome with another insult to Jesus!” Leo I saw you comment about not being able to contact David Icke , I think I know why . I was turned off Icke when he put on his site about Tommy Robinson being beaten up in Prison according to a fake BBC report , he has totally ignored the impact that Islam Is having on the West yet rails against the Jews , Trump , ect . He is either a closet Islamist or too frightened to say anything about it . Moral Relativism is a CANCER. Period.
\item Michael Piggin, an 18-year-old British teenager armed with guns and explosives, planned a Columbine-style massacre in Loughborough, Leicestershire, an Old Bailey jury heard Monday. Piggin from Loughborough, Leics was arrested February 2013 on two terrorism offences. Two other 18-year-olds who were arrested along with Piggin admitted to possessing petrol and pipe bombs. The trial is expected to last six to eight weeks. Piggin, who suffers from Asperger syndrome, scribbled Nazi slogans and had a Swastika flag on his wall. He also drew a cartoon detailing in which knives, a gun and bombs were concealed in a trench coat. The 'worst and most life-changing things' happened at school and council officials had 'tortured my family over the years,' Piggin wrote in a Che Guevara notebook adorned with Nazi swastikas and English Defence League (EDL) mottos, the Guardian reports. He also wrote: "There's too many Muslims in the UK, we must stop them coming over here. Send the ones already here back home before they take over," Mirror reports. "I'm ready to die for my cause," Piggin said over a dictaphone. Calling Piggin a terrorist, Max Hill QC said that he has extreme views and was serious about executing his plans. The 1999 Columbine High School massacre involved two senior students Dylan Klebold and Eric, who fatally shot 12 students and one teacher. The school shooting also left 24 other students injured while attempting to flee the school. The pair then committed suicide. The massacre conspiracy also made references to fanatic Anders Breivik's terrorist attacks in Norway. Breivik killed 77 people in Oslo and on Utoya island 2011. He also made a fascist salute at his trial before being jailed for 21 years.
\item A minds of children. The U.S. Department of Justice agrees. Last March the Home School Legal Defense Fund (HSLDF), the organization defending the Romeikes, created a White House petition asking President Obama to protect the family’s asylum. Obama declined to respond, but his Department of Justice sure went on the offensive. In its June 26, 2013, brief to the U.S. Sixth Circuit Court of Appeals, Obama’s DOJ essentially argued that the German government is right to take children away from their parents and force them into government schools because “teaching tolerance to children of all backgrounds helps to develop the ability to interact as a fully functioning citizen of Germany.” Tolerance, yes. Tolerance of the state, not tolerance by the state. We can be sure that the DoJ has the same view of public schools in the USA, but it knows it cannot enforce this view here. The DoJ did not reverse itself. The DHS did. Continue Reading on Wonder how long it is going to be before the 'State' outlaws that here? What Gary shows is that the DOJ gets sweaty palms just ANTICIPATING more statism (enslavement) of education and the family. It uses “tolerance” to implement intolerance! Wow, what an indictment of our state school system!! 1) The Germans are the ones we got the idea of "public" schools from in the first place. It's a masterful scheme for indoctrination 2) By this ruling, the DHS even admits this 3) In its statement, the DOJ and Obama administration come right out and claim that the purpose of public education is to indoctrinate students in social values the PTB regards as important–in this case, "tolerance". Implication: Homeschoolers are not being taught "tolerance", and maybe are even being taught intolerance!! Whew. The DOJ is incorrect. It is, under 'the high holder', actually the DOIJ or the department of injustice. That would require the Supreme Court overturning a whole string of cases involving people from a whole bunch of different religions. At different times such diverse religious bodies as the Roman Catholics, the Lutheran Church-Missouri Synod, the Amish and the Jehovah's Witnesses have gone to court and won. Any effort to ban it would be almost impossible. Holder is the beholder of injustice, lawlessness, and treason. WEBSTER’S DICTIONARY (1967) defines TREASON as “any attempt to overthrow the government of a state to which one owes allegiance”. No worries. It'll all come out in the war–Civil War II, the second American Revolution, the war to restore the Republic. Call it whatever you want. Let's just get on with it. just a guess – some mid level bureaucrat at the DHS who hasn’t yet had all the humanity crushed out of them yet took pity on the family. Good. We have to make sure that nobody changes their minds. The aim is indoctrination, to create internal controls that support the State. Everything else is secondary. This is really good published article. Such a great yet interesting post. Thank you very much for sharing this nice stuff.
\item PML-N Vice President Maryam Nawaz on Sunday, in a cryptic message, said that “anyone who speaks against Nawaz Sharif will have their tongues ripped out”. During a youth convention in Lahore the PML-N leader also declared that it is “now time to stop giving sacrifices and demand accountability” from the government. The PML-N vice president said that it would be said in the past that only Sindh, Balochistan and Khyber Pakhtunkhwa “stand up to the selectors” but now “Punjab has also stood up” to them.
\item If you want a view of the future just look at Vancouver.  Vancouver is a complete mess with that very small minority of Chinese moving to other countries.  That small minority of people are making a huge mess.  It doesn't matter there are a billion Chinese in the world the tiny amount moving about and destroying other economies and causing this disdain is the issue.  They have no intention of assimilating into local cultures or contributing to the country they are moving into.  Italy had to hire Chinese speaking police to handle the neuvo or newly rich Chinese tourist because they simply do not care or want to act civilly when they visit out countries.  Rant off.
\item A Fulani group recognized as Fulani Nationality Motion, FUNAM, has claimed accountability for the assault on Governor Samuel Ortom of Benue State. FUNAM mentioned it deliberate to kill Ortom on Saturday, but it surely’s plan suffered a setback attributable to a slight communication error. A press release by one Umar Amir Shehu mentioned Ortom stood towards open grazing and Fulani curiosity within the nation. The assertion entitled: “Why we attacked Ortom” reads: “Our consideration has been drawn to media studies at present speculating about who attacked the Governor of Benue State, Samuel Ortom. “Sure. Sure, We did. The Fulani Nationality Motion, (FUNAM) carried out the assault. We’ve got real causes. We acted on behalf of Thousands and thousands of Fulani folks in 15 nations. “It’s a case of vengeance towards an infidel who has used his money and time, deployed in destroying the Fulani values and inheritance. “Our brave fighters carried out this historic assault to ship a terrific message to Ortum and his collaborators: The place ever you’re, as soon as you’re towards Fulani long run curiosity, we will get you down. This can be a clear warning. We hope those that take us with no consideration will get the indeniable message. “Our intention is unequivocal: TO KILL HIM. That mission will someday be fulfilled and really quickly too.” “Eleven of FUNAM operatives have been concerned within the assault. Ortom escaped at present due to a slight technical communication error. Nest time, he is not going to be fortunate. We are able to guarantee him and his supporters.” On Saturday, herdsmen had attacked Ortom on his farm at Tyo-mu village, alongside Makurdi – Gboko street. Reacting, the governor had alleged that linked the assassination try and Miyetti Allah Cattle Breeders Affiliation of Nigeria, MACBAN. The governor claimed that MACBAN had concluded plans to assassinate him over his place on open grazing. Ortom alleged that the Myetti Allah took the choice to get rid of him throughout a gathering they organised lately in Yola. He famous that his life is within the fingers of the Supreme God and no Fulani man can kill him, mentioned he obtained intelligence that the Affiliation deliberate to kill his safety aides and subsequently seize him alive.
\item So let Kit Siang and DAP be warned. The English word amokcomes from the Malay word mengamuk. And that’s because mengamuk is only in the Malay DNA. And when the Malays do mengamuk not even their own mother can stop them. So let Kit Siang and DAP be warned. THE CORRIDORS OF POWER Raja Petra Kamarudin Dedak-eater Lim Kit Siang and his secular-republican party, DAP, is pushing Malaysia to the brink of a civil war. If you think the many racial and religious conflicts that occurred since 2nd September 1945 till 13th May 1969 are tragedies, wait till you see this next one. In the two-week Chinese riot of 2nd September 1945 and the many other racial and religious conflicts that occurred thereafter — the ‘big one’ on 13th May 1969 — they were isolated and contained affairs. It was either in Johor, Penang, Bukit Mertajam, Kuala Lumpur, Singapore, etc. Even in the 1940 riot when the Chinese Communists slaughtered Malays and British — men, women and children — it was only in Singapore and parts of Southern Johor but not nationwide. Secondly, they were single-issue affairs. They were either organised by the Communists (such as the 1940 riot in Singapore and the 2nd September 1945 riot), or about religion (such as the riots in Johor, Penang and Singapore), or about race (such as the riots in Kuala Lumpur and Bukit Mertajam). This next one, which is going to be more like a cvil war, is going to be about a multitude of issues, race and religion being only two of the many reasons. When the conflict is about hunger, poverty, race, religion, or a misunderstanding (such as a temple ringing a bell when Muslims are praying in the mosque next door, or a noisy wedding next door to a funeral) they are easier to solve. But when the conflict is about hatred, then it becomes harder to end the conflict. There are many examples over just the last 100 years alone (no need to research the last 1,000 years) where conflicts erupted due to pent-up hatred over decades or even centuries. Finally it explodes and when it does it can only be resolved through either ethnic cleansing or partition. Closer to home we have Burma, Thailand and Philippines, which are are still suffering from conflicts, Burma being the worst. Indonesia launched ethnic cleansing in 1965 and an estimated three million Chinese and Communists were massacred. Indonesia solved the problem by literally killing the problem. The conflict in Burma is actually not new but started at the time of WWII during the Japanese occupation due to animosity between the Rohingya pendatang and Burmese Bumiputeras. So Burma is a 70-year old problem that has now turned to ethnic cleansing. Eventually misunderstandings turn to hatred and hatred turns to bloodshed. It is the normal progression of events that 3,500 years of recorded history has proven. And in the end a trigger sets it off and the blood starts to flow. The last two years have not been good for race relations and religious tolerance. Kit Siang and DAP have been testing the patience of the Malays and Muslims. It is almost like Kit Siang and DAP are openly declaring war on the Malays, Islam and the Raja-Raja Melayu. This provocation cannot go on forever. Sooner rather than later something is going to break. And when it does Kit Siang will run away and hide, like he did in May 1969, and leave innocent Malaysians to get slaughtered in the riots that are going to spread throughout Malaysia..
\item However, what I object to is the idea that these people we are supposed to hate  "Supposed to hate" what fiction is this? I'm not proposing this. Maybe the BNP are but that's a tiny perspective. There is a great deal of nuance between the BNP position and yours, you should explore it.   \&gt;However, what I object to is the idea that these people we are supposed to hate are special  Special? What do you mean by this?   \&gt; I've known plenty of men who've oppressed women, had violent tendencies and supported appalling politics, white as the driven snow.  No one is disputing "woman oppresses", people with "violent tendencies" or those who support "appalling politics" cannot be white. Whats the point of these boring strawmen?   \&gt;But Muslims are especially bad, for some reason, we can't tolerate them, despite their supposedly incompatible views are only about 40 years behind the trend in our society. Even though the features supposedly so unique and "unwestern" are on display all around us. It rather makes me think that the distinguishing thing is that ultimately they tend to look a bit different and and we're rationalising bullshit.  Just look at the Muslim world. Things we take for granted in Britain are non-existent in many of those countries. The state and Islam are basically entwined (ours is, but only ceremonially these days), Sharia law is the norm (although it differs from Saudi's ultra rigid following to more liberal states like Algeria). Religious difference is not tolerated, women genuinely are 2nd class systems (and no we didn't have these beliefs in 1976 m8). All these things derive from Islam and it's followers.   ...and just look at the appalling record of Muslim integration in Britain. They stick to their own; in their own streets of Bradford or London. Have a laissez faire attitude to raping children (well, white ones anyway), very rarely will "marry" with other faiths (and if they do, the non-Muslim is converting to Islam as a pre-requisite), they've been guilty of infiltrating schools and indoctrinating children (i reckon we're on the tip of the iceberg with that one), furthermore repeated surveys demonstrating attitudes are at odds with everybody else whether it's sharia law or how many sympathise with the motivations of terrorists....I mean, we want more of this? Is that what you want? If you had to picture Britain in 50 years time, you want more Islam? You want more religiosity?
\item St. Paul in his letter to the Kligontians. Would the crowds have followed Jesus if he looked like this? He couldn't get a miracle passed if he had this guy's kisser. One of the great misconceptions about Jesus is that he was Jewish. This is of course ignorance in the extreme - Jesus Christ was a Christian. All you have to do is read about him, because if the Bible said he was Jewish it's news to me. And none of the stuff about Jewish things ever makes it into the movies, so it didn't happen. In fact, researchers in Hollywood have proven beyond a reason or a doubt that Jesus was one of the founders of Christianity, along with Moses and Cleopatra and maybe even John that Baptist. Jesus was (as we all know, from the movies) murdered by the Jews due to his religious beliefs and good looks, and you can tell from the better ones that the Jews really had some kind of bone to pick with the guy. All of that proves that Jesus was not Jewish. People have to stop believing this un-movielike lie. Here are several reasons why I know what I know. Jesus never wore glasses, or had affairs with my mother which caused my dad to leave me and my older brother. Never. And Jesus never ate babies, or cut off his genitalia (why would he?), and he never stole money from the his followers or mom's rainy-day account when he could help it. In fact he was highly critical of such things, especially when he ran the merchants out of the temple for not giving him the correct change. Jesus never celebrated Hanukkah or Passover or Christmas, nor any of those other bogus holidays promoted in the Jewish bestseller, the Book of Leviticus. He especially did not celebrate the Year of Jubilee. Jesus instead promoted a philosophy totally at odds with the ignorant Jewish vegetarian rule of "Thou shalt not kill", or the idol-worshipping ways of orthodox Judaism, or even the things Arabs do. I doubt if Jesus even came from the Middle-East at all. And Jesus wasn't like any of those heathen virgin birth Sumerian, Egyptian, Roman and Persian Gods who were born on December 25th, not in any way. Because he was a Christian, and proud of it. A balanced and fair depiction of your typical historical Jewish activity. Jesus would never cook humans, not in one million years. Jesus was the son of God. God is a Christian, and Mary was also a Christian long before she agreed to carry Jesus around inside her and then let him loose into the world. So there you go, ipso facto. If God was Jewish what do you know? His mother's looks, fair-skinned, takes pride in his appearance - blue eyed, blonde, and gentle. Compare him to typical Jewish otherworldly DNA and it's clear that Jesus was not even part Jewish. The Jews killed Jesus for his religious viewpoints and maybe his hair. If Jesus were a Jew, which he never was and neither were his parents or his relatives or his Rabbi, the Jews would not have done this, as Jews do not kill their own kind. Much like their ancestors, the ants, they are a hive-mind, and to kill one of their own kind would cause a chain-reaction like lemmings walking up to the edge of a cliff and deciding to jump off. Jesus was the lemming who said "Let's don't jump this time, let's do something else now". Since the Jews killed Jesus with a big piece of wood, like a vampire only on the outside, it is safe to assume he most certainly wasn't Jewish but Christian. As a Christian he went to heaven, which he would never have found if he was Jewish because everyone knows and the movies tell us that the Jews have a different kind of heaven, one sort of like that E.T. film where the alien guy points to his heart and goes home or something. If Jesus or my dad had been Jewish and real strong the story would have ended more or less like this. But Jesus was Christian, and so he knew he'd be in heaven with his dog soon enough. Jesus told people to be Christian and watch out for Jewish culture and food, not out of hate or fear or indigestion, but out of a divine purpose and an instinct for protection. God, as stated and proven above, is Christian, and thus could no longer stand to see the Jews taking over things like the film industry, so he sent Jesus down to earth to show the Jews how it should properly be done (psst. Revenue sharing versus a piece of the percentage). "Who're you calling a Jew? You and what Roman army?" Mary Magdelene was Christian all the way, not Jewish. When Jesus started to date her he made sure she was Christian. She even knew some of the apostles, who were all Christian and not Jewish. One of the apostles, John, used to date her before Jesus did, and they had a man-thing going for awhile about who was dating her now, and back off bro I saw her first, and I dated her before you did so she's mine, and uh-uh, O no you don't, and then Jesus won her heart because you know why? He was Christian! That's why. Jesus' name on his birth certificate is "Jesus Christ", not "Jesus Jew" or "Jesus, King of the Jews". If he was Jewish he wouldn't have been named "Christ" right from the get-go, now would he? Names mean things, and to be named Jesus Christ was pretty cool for the guy, who could have been named most anything. It's pretty simple to say that this alone proves he was Christian. The Father, Son, and Holy Ghost all dance to celebrate Jesus' Christianhood. So there! ↑ Ironically, his followers have now set up mass markets to sell all kinds of Christian symbols. Go figure. This page was last edited on 9 December 2014, at 00:47.
\end{enumerate}

\end{document}